\documentclass[12pt]{article}

\ifx\pdfpageattr\undefined
\else
  \pdfpageattr{/Rotate 0}
\fi
\usepackage[utf8]{inputenc}
\usepackage[super,comma,sort&compress]{natbib} 
\usepackage[rmargin=.75in,lmargin=.75in,tmargin=1in,bmargin=1in]{geometry}
\usepackage{graphicx}
\usepackage{amsmath,amssymb,amsfonts,amsthm}
\usepackage{xcolor}
\usepackage{booktabs}
\usepackage{latexsym}
\usepackage{xfrac}
\usepackage{tcolorbox}
\usepackage{tikz}
\usepackage{float}
\usepackage{microtype}
\usepackage{pdflscape}
\usepackage{url}
\usepackage[T1]{fontenc}
\usepackage{caption}
\usepackage{subcaption}

\newcommand{\key}{\textbf}

\newcommand{\bluesquare}{\textcolor{blue}{\rule{.9em}{.9em}}}

\begin{document} 

    \title{Linguistic Structure from a Bottleneck on Sequential Information Processing}


\author{Richard Futrell$^1$ and Michael Hahn$^2$ \\
$^1$ University of California, Irvine \\
$^2$ Saarland University
}

\date{}

\maketitle

\abstract{
Human language has a distinct systematic structure, where utterances break into individually meaningful words which are combined to form phrases. 
We show that natural-language-like systematicity arises in codes that are constrained by a statistical measure of complexity called predictive information, also known as excess entropy. Predictive information is the mutual information between the past and future of a stochastic process. 
In simulations, we find that such codes break messages into groups of approximately independent features which are expressed systematically and locally, corresponding to words and phrases.
Next, drawing on crosslinguistic text corpora, we find that actual human languages are structured in a way that reduces predictive information compared to baselines at the levels of phonology, morphology, syntax, and lexical semantics.
Our results establish a link between the statistical and algebraic structure of language and reinforce the idea that these structures are shaped by communication under general cognitive constraints.
}

\newpage

\section*{Introduction}
\begin{figure}
\centering
\includegraphics[width=\textwidth]{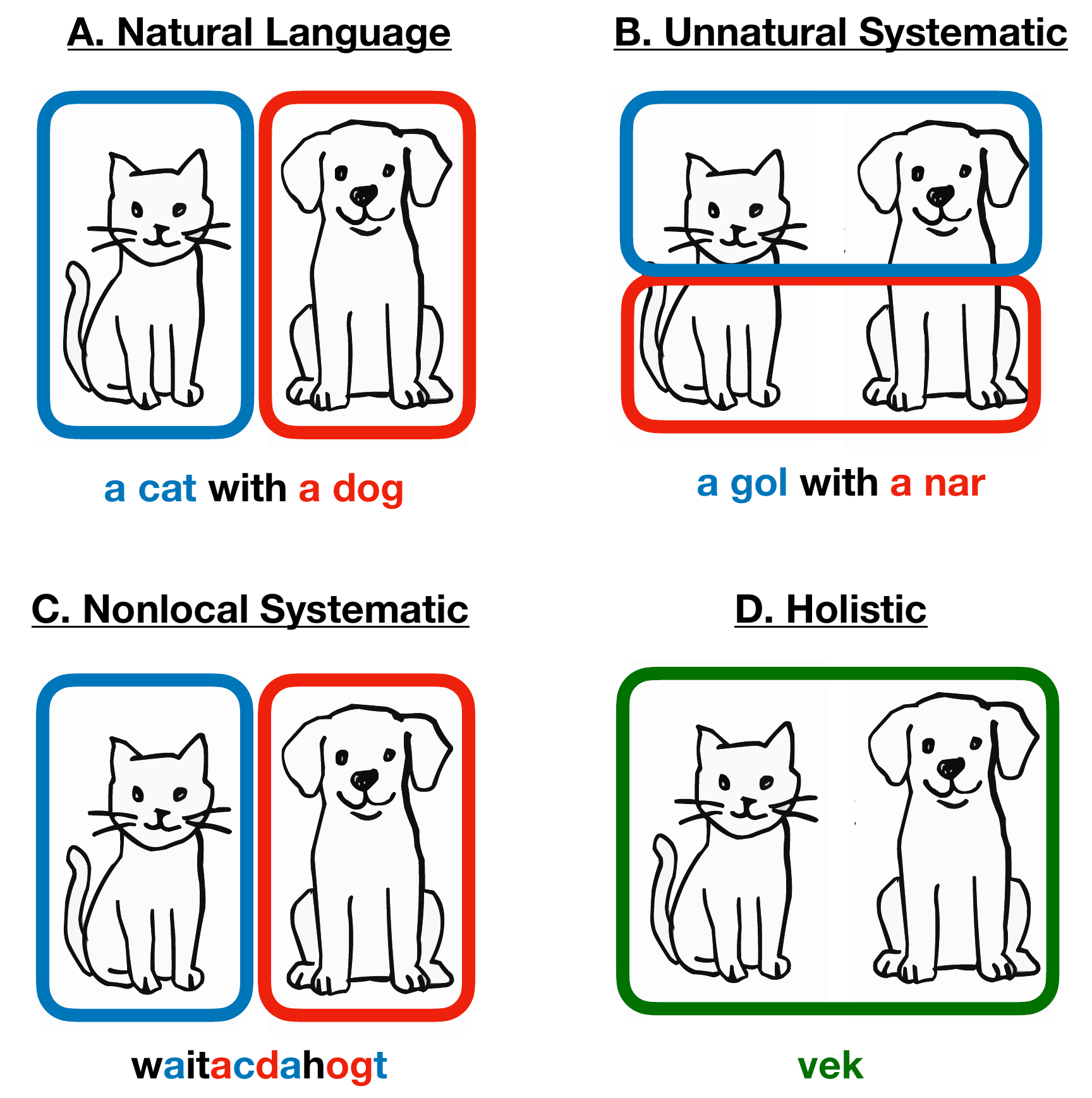}
\caption{Example utterances describing an image in English and various hypothetical languages. \textbf{A.} An English utterance exhibiting systematicity and locality. \textbf{B.} An unnatural systematic language in which \emph{gol} means a cat head paired with a dog head and \emph{nar} means a cat body paired with a dog body. \textbf{C.} A nonlocal but systematic language in which an utterance is formed by interleaving the words for `cat' and `dog'. \textbf{D.} A holistic language in which the form \emph{vek} means `a cat with a dog' with no correspondence between parts of form and parts of meaning.}
\label{fig:systematic}
\end{figure}

Human language is organized around a systematic, compositional correspondence between the structure of utterances and the structure of the meanings that they express\cite{frege1923gedankengefuege}. For example, an English speaker will describe an image such as Fig.~\ref{fig:systematic}A with an utterance like \emph{a cat with a dog}, in which the parts of the the image correspond regularly with parts of the utterance such as \emph{cat}---what we call words. This way of relating form and meaning may seem natural, but it is not logically necessary. For example, Fig.~\ref{fig:systematic}B shows an utterance in a hypothetical counterfactual language where meaning is decomposed in a way which most people would find unnatural: here we have a word \emph{gol} which refers to a cat head and a dog head together, and another word \emph{nar} which refers to a cat body and a dog body together. Similarly, Fig.~\ref{fig:systematic}C presents a hypothetical language which is systematic but with an unnatural way of decomposing the utterance: here the utterance contains individually meaningful subsequences \emph{a cat}, \emph{with}, and \emph{a dog}, but these are interleaved together, rather than concatenated as they are in English. We can even conceive of languages such as in Fig.~\ref{fig:systematic}D, where each meaning is expressed holistically as a single unanalyzable form\citep{jespersen1922language,wray1998protolanguage}---in fact, this lack of systematic structure is expected in optimal codes like Huffman codes\citep{huffman1952method,futrell2022information}. Why is human language the way it is, and not like these counterfactuals?

We argue that the particular structure of human language can be derived from general constraints on sequential information processing.
We start from three observations: 
\begin{enumerate}
\item Utterances consist, to a first approximation, of one-dimensional sequences of discrete symbols (for example, phonemes).
\item The ease of production and comprehension of these utterances is influenced by the sequential predictability of these symbols down to the smallest timescales\cite{goldmaneisler1957speech,ferreira2002incremental,bell2009predictability,smith2013effect,heilbron2022hierarchy,ryskin2023prediction}.
\item Humans have limited cognitive resources for use in sequential prediction\cite{miller1963finitary,bratman2010new,christiansen2016nowornever,futrell2020lossy,ferdinand2024humans}.
\end{enumerate}
Thus we posit that language is structured in a way that minimizes the complexity of sequential prediction, as measured using a quantity called predictive information: the amount of information about the past of a sequence that any predictor must use to predict its future\cite{grassberger1986toward,bialek2001predictability}, also called excess entropy\citep{crutchfield2003regularities,dkebowski2020information}. Below, we find that codes which are constrained to have low predictive information within signals have systematic structure resembling natural language, and we provide massively cross-linguistic empirical evidence based on large text corpora showing that natural language has lower predictive information than would be expected if it had different kinds of structure.

\begin{figure}
\centering
\includegraphics[width=.5\textwidth]{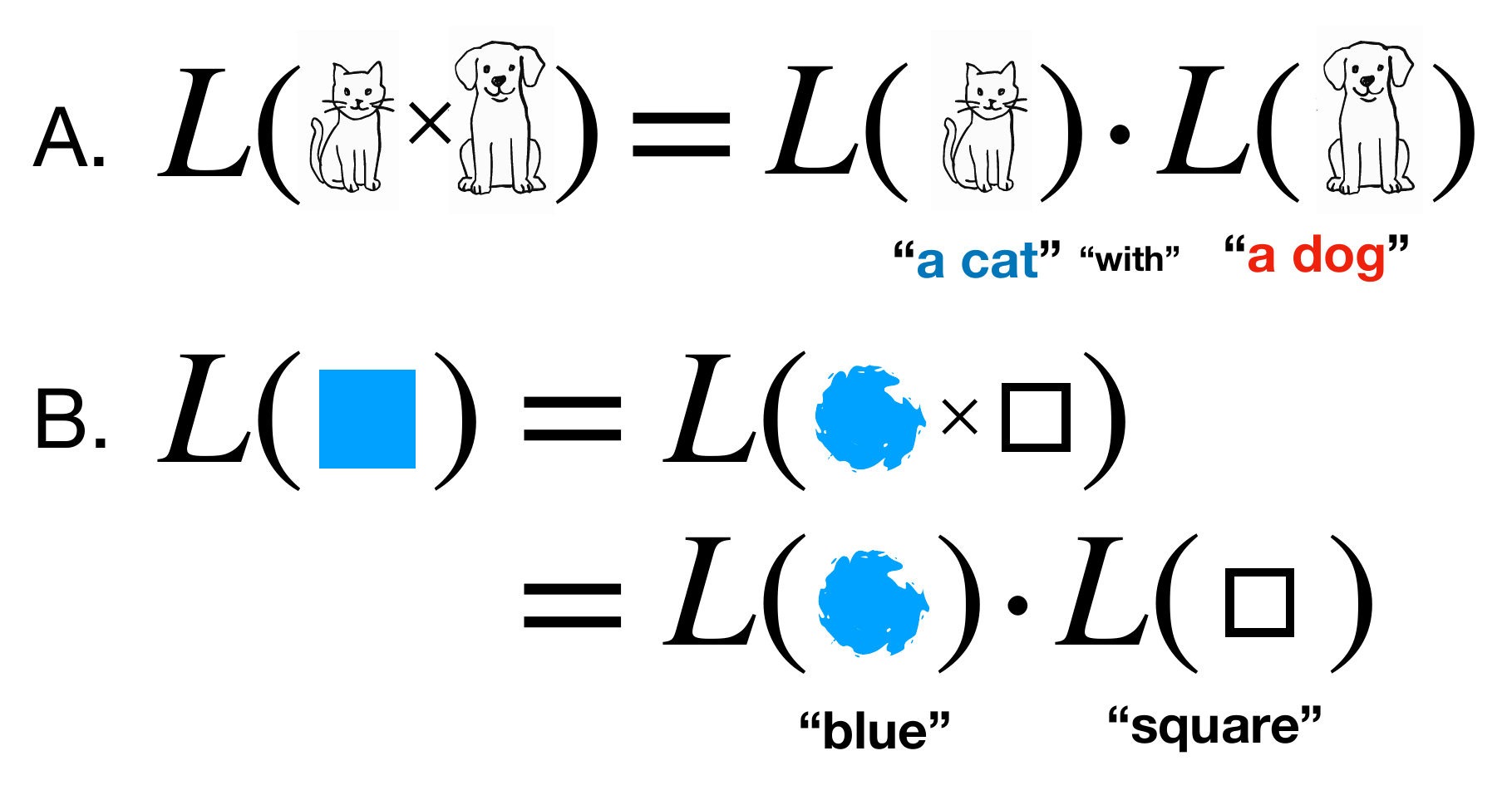}
\caption{Two examples of linguistic systematicity as a homomorphism. $L(\cdot)$ stands for the English language, seen as a function from meanings to forms (strings). \textbf{A.} The meaning naturally decomposes into two features corresponding to the two animals. The form \emph{a cat with a dog} decomposes systematically into forms for the cat and the dog, concatenated together with the string \emph{with} between them. \textbf{B.} The meaning naturally decomposes into two features, corresponding to color and shape. The form \emph{blue square} decomposes systematically into forms for the color and the shape, concatenated together.}
\label{fig:homomorphism}
\end{figure}

\section*{Results}

\subsection*{Explananda}
First we clarify what we want to explain. Taking a maximally general stance, we think of a language as a function mapping meanings to forms, where meanings are any objects in a set $\mathcal{M}$, and forms are strings drawn from a finite alphabet of letters $\Sigma$, typically standing for phonemes. We say a language is \key{systematic} when it is a homomorphism\cite{montague1970universal,janssen1997compositionality}, as illustrated in Fig.~\ref{fig:homomorphism}. That is, if a meaning $m$ can be decomposed into parts (say $m = m_1 \times m_2$), then the string for that meaning decomposes in the same way:
\begin{equation}
\label{eq:systematicity}
L(m_1 \times m_2) = L(m_1) \cdot L(m_2),
\end{equation}
where $\cdot$ is some means of combining two strings, such as concatenation.
For example, an object \bluesquare~would be described in English as $L(\bluesquare) = \text{\emph{blue square}}$. The meaning \bluesquare~is decomposed into features for color and shape, and these features are expressed systematically as the words \emph{blue} and \emph{square} concatenated together. 

We wish to explain why human languages are systematic, and furthermore why they decompose meanings in the way they do, and why they combine strings in the way they do. In particular, meanings are decomposed in a way that seems natural to humans (that is, like Fig.~\ref{fig:systematic}A and not Fig.~\ref{fig:systematic}B), a property we call \key{naturalness}. Also, strings are usually combined by concatenation (that is, like Fig.~\ref{fig:systematic}A and not like Fig.~\ref{fig:systematic}C), or more generally by some process that keeps relevant parts of the string relatively close together. We call this property \key{locality}. 

Influential accounts have held that human language is systematic because language learners need to generalize to produce forms for never-before-seen meanings\cite{kirby1999syntax,smith2003complex,franke2014creative,kirby2015compression}. Such accounts successfully motivate systematicity in the \emph{abstract} sense, but on their own they do not explain naturalness and locality.
However, a theory of systematicity must have something to say about these properties, because if we are free to choose any arbitrary functions $\times$ and $\cdot$, then any function $L$ can be considered systematic in the sense of Eq.~\ref{eq:systematicity}, and the idea of systematicity becomes vacuous\cite{zadrozny1994compositional}.

In existing work, naturalness and locality are explained via (implicit or explicit) inductive biases built into language learners\cite{batali1998computational,kirby1999syntax,ke2006language,tria2012naming,lazaridou2017multiagent,mordatch2017emergence,steinertthrelkeld2020toward,kucinski2021catalytic,begus2024basic} or stipulations about the mental representation or perception of meanings\cite{nowak2000evolution,barrett2007dynamic,franke2016evolution,barrett2020evolution,culbertson2020world}.
In contrast, we aim to explain natural local systematicity in language from maximally general principles, without any assumptions about the mental representation of meaning, and with extremely minimal assumptions about the structure of forms---only that they are ultimately expressed as one-dimensional sequences of discrete symbols.

\subsection*{Predictive Information}

We measure the complexity of sequential prediction using \key{predictive information}, which is the amount of information that any predictor must use about the past of a stochastic process to predict its future. (Below, we assume familiarity with information-theoretic quantities of entropy and mutual information\cite{cover2006elements}.) 
Given a stationary stochastic process generating a stream of symbols $\dots, X_{t-1}, X_t, X_{t+1}, \dots$, we split it into `the past' $X_{\text{past}}$, representing all symbols up to time $t$, and `the future' $X_{\text{future}}$, representing all symbols at time $t$ or after. The predictive information or excess entropy\cite{bialek2001predictability,crutchfield2003regularities} $E$ is the mutual information between the past and the future:
\begin{equation}
E = \mathrm{I}[X_{\text{past}} : X_{\text{future}}].
\end{equation}
We calculate the predictive information of a language $L$ as the predictive information of the stream of letters generated by repeatedly sampling meanings $m \in \mathcal{M}$ from a source distribution, translating them to strings as $s = L(m)$, and concatenating them with a delimiter in between.

\begin{figure}
\centering
\includegraphics[width=.5\textwidth]{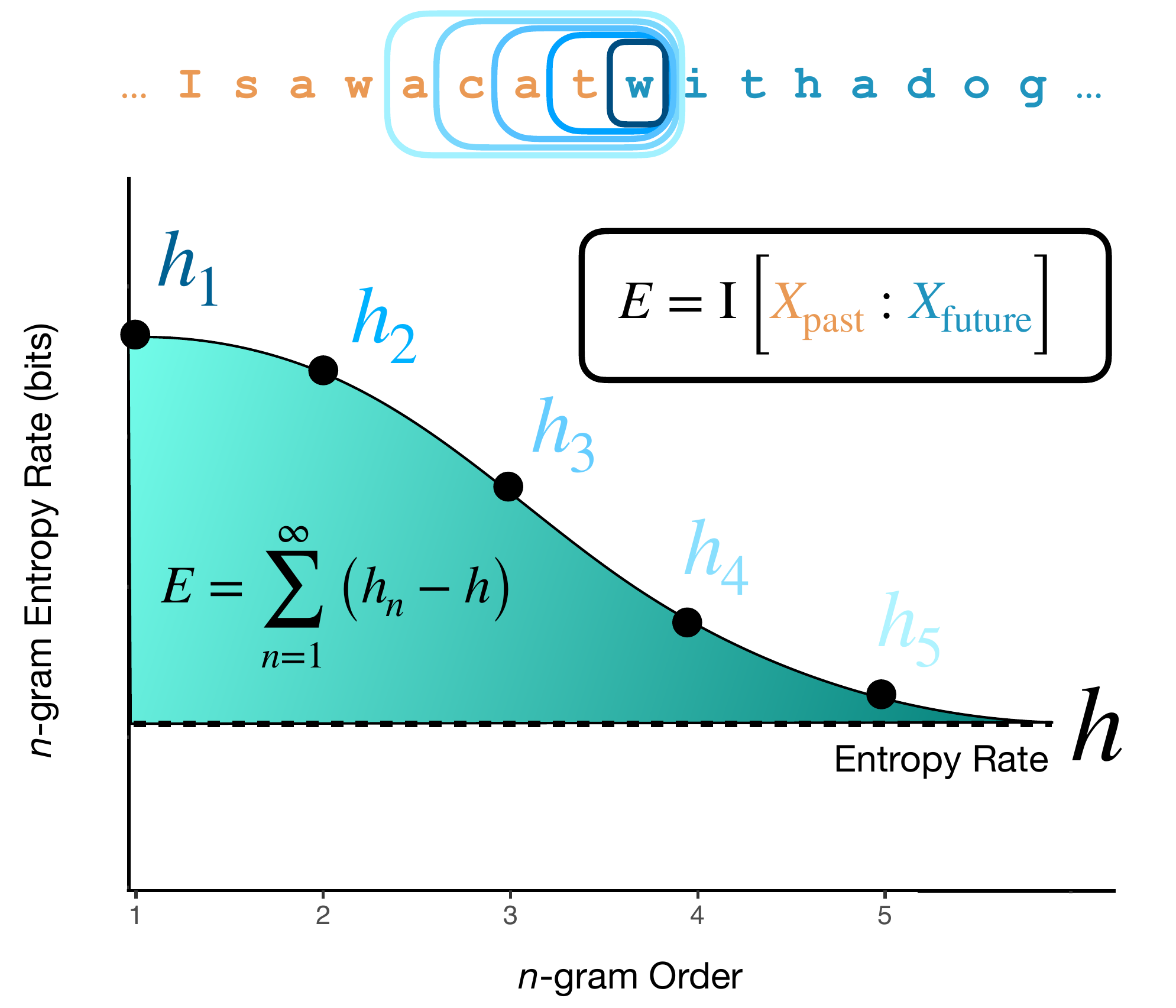}
\caption{Schematic calculation of predictive information as the sum of $n$-gram entropies $h_n$ minus the asymptotic entropy rate $h$.}
\label{fig:ee-calc}
\end{figure}

Predictive information can be calculated in a simple way that gives intuition about its behavior. Let $h_n$ represent the \key{$n$-gram entropy} of a process, that is, the average entropy of a symbol given a window of $n-1$ previous symbols:
\begin{equation}
h_n = \mathrm{H}[X_t \mid X_{t-n+1}, \dots, X_{t-1}].
\end{equation}
As the window size increases, the $n$-gram entropy decreases to an asymptotic value called the \key{entropy rate} $h$. The predictive information represents the convergence to the entropy rate,
\begin{equation}
\label{eq:entropy-rate-sum}
E = \sum_{n=1}^\infty \left(h_n - h\right),
\end{equation}
as illustrated in Fig.~\ref{fig:ee-calc}. This calculation reveals that predictive information is low when symbols can be predicted accurately based on local contexts, that is, when $h_n$ is close to $h$ for small $n$.

\begin{figure}
\centering
\includegraphics[width=1.05\textwidth]{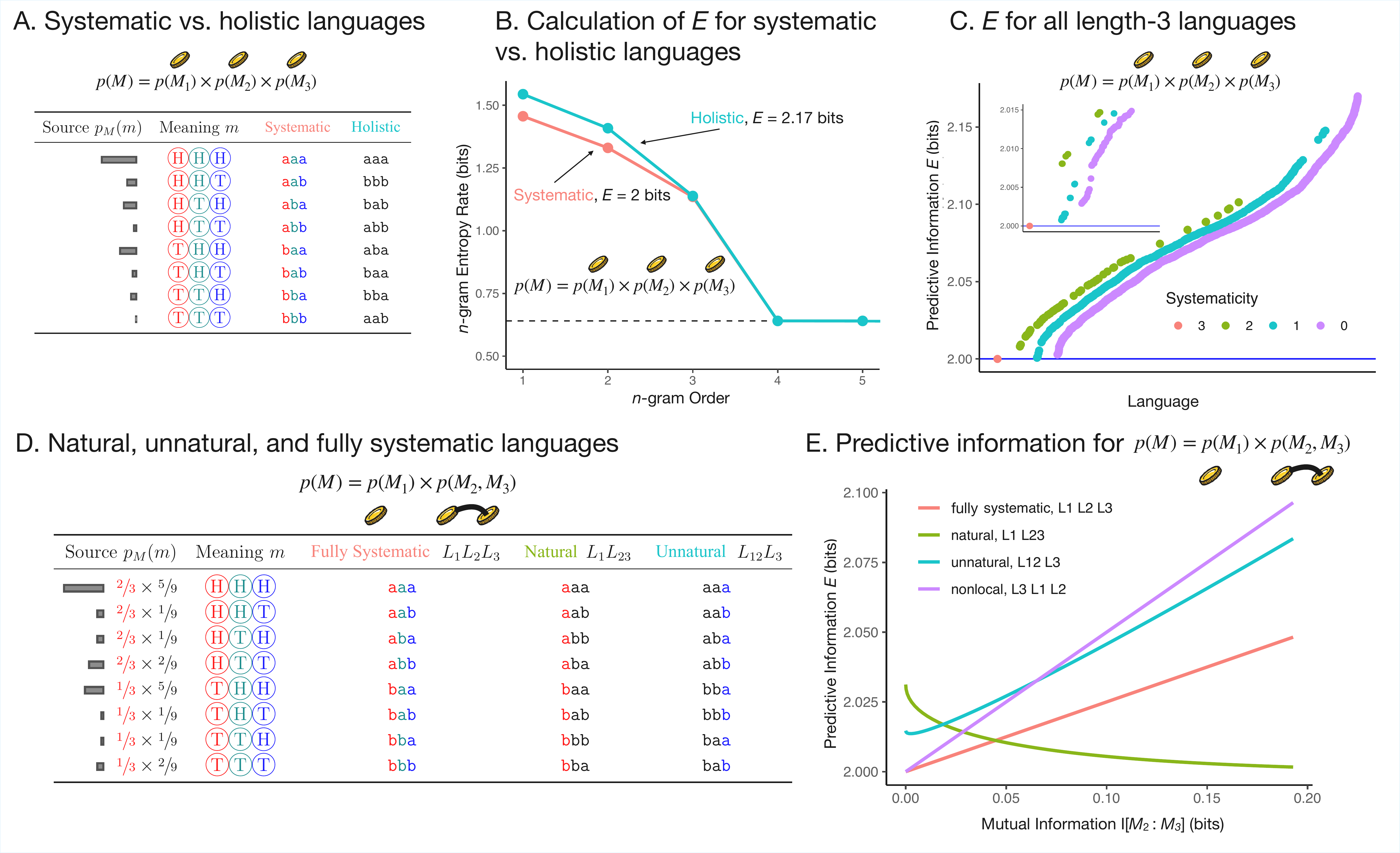}
\caption{Simulations of languages for coinflip distributions. 
\textbf{A.} Two unambiguous languages for meanings consisting of three weighted coinflips. In the systematic language, each letter corresponds to the outcome from one coinflip. In the holistic language, there is no natural systematic relationship between the form and the meaning. 
\textbf{B.} Calculation of predictive information for the source and two languages in panel A. The systematic language has lower predictive information.
\textbf{C.} Predictive information of all bijective mappings from meanings to length-3 binary strings, for the meanings and source in panel A. Languages are ordered by predictive information and colored by the number of coinflips expressed systematically: 3 for a fully systematic language and 0 for a fully holistic language. The inset box zooms in on the low predictive information region. 
\textbf{D.} Languages used in panel E along with an example source, which has mutual information $\operatorname{I}[M_2 : M_3] \approx 0.18$ bits.
\textbf{E.} Predictive information of various languages for varying levels of mutual information between coinflips $M_2$ and $M_3$ (see text). Zero mutual information corresponds to panels B and C. The `natural' language expresses $M_2$ and $M_3$ together holistically. The `unnatural' language expresses $M_1$ and $M_2$ together holistically.}
\label{fig:coin-sims}
\end{figure}

\subsection*{Simulations}

The following simulations show that when languages minimize predictive information, they express approximately independent features systematically and locally in a way that corresponds to words and phrases in natural language.

\paragraph{Systematic expression of independent features} Consider a set of meanings consisting of the outcomes of three weighted coinflips. In a natural systematic language, we would expect each string to have contiguous `words' corresponding to the outcome of each individual coin, whereas a holistic language would have no such structure, as shown in the examples in Fig.~\ref{fig:coin-sims}A. It turns out that, for these example languages, the natural systematic one has lower predictive information, as shown in Fig.~\ref{fig:coin-sims}B. In fact, among all possible unambiguous length-3 binary languages, predictive information is minimized in all and only the systematic languages, as shown in Fig.~\ref{fig:coin-sims}C. 

Intuitively, the reason systematic languages minimize predictive information here is that the features of meaning expressed in each individual letter are independent of each other, and so there is no statistical dependence among letters in the string. The general pattern is that an unambiguous language that minimizes predictive information will find features that have minimal mutual information and express them systematically. See SI~Appendix~A for formal arguments to this effect.

\paragraph{Holistic expression of correlated components} What happens to predictive information when the source distribution cannot be expressed in terms of fully independent features? In that case, it is better to express the more correlated features holistically, \emph{without} systematic structure. This holistic mapping is what we find in natural language for individual words (or more precisely, morphemes), according to the principle of arbitrariness of the sign\cite{saussure1916cours}. For example, the word \emph{cat} has no identifiable parts that systematically correspond to features of its meaning. Furthermore, as we will discuss below, morphemes in language typically encode categories whose semantic features are highly correlated with each other\citep{rosch1978principles}.

We demonstrate this effect by varying the coinflip scenario above. Denote the three coinflips as $M_1$, $M_2$, and $M_3$. Imagine the second and third coins and are tied together, so that their outcomes $M_2$ and $M_3$ are correlated, as in the example in Fig.~\ref{fig:coin-sims}D. In the limit where $M_2$ and $M_3$ are fully correlated, these coinflips have effectively become one feature. Fig.~\ref{fig:coin-sims}E shows predictive information for a number of possible languages in this setting, as a function of the mutual information between the tied coinflips $M_2$ and $M_3$. In the low-mutual-information regime---where $M_2$ and $M_3$ are nearly independent---the best language is still fully systematic. However, as mutual information increases, the best language is one that expresses the tied coinflips $M_2$ and $M_3$ together holistically, as a single `word'. An unnatural language that expresses the uncorrelated coinflips $M_1$ and $M_2$ holistically is much worse, as is a nonlocal systematic language that breaks up the `word' corresponding to the correlated coinflips $M_2$ and $M_3$.

\begin{figure}
\centering
\includegraphics[width=\textwidth]{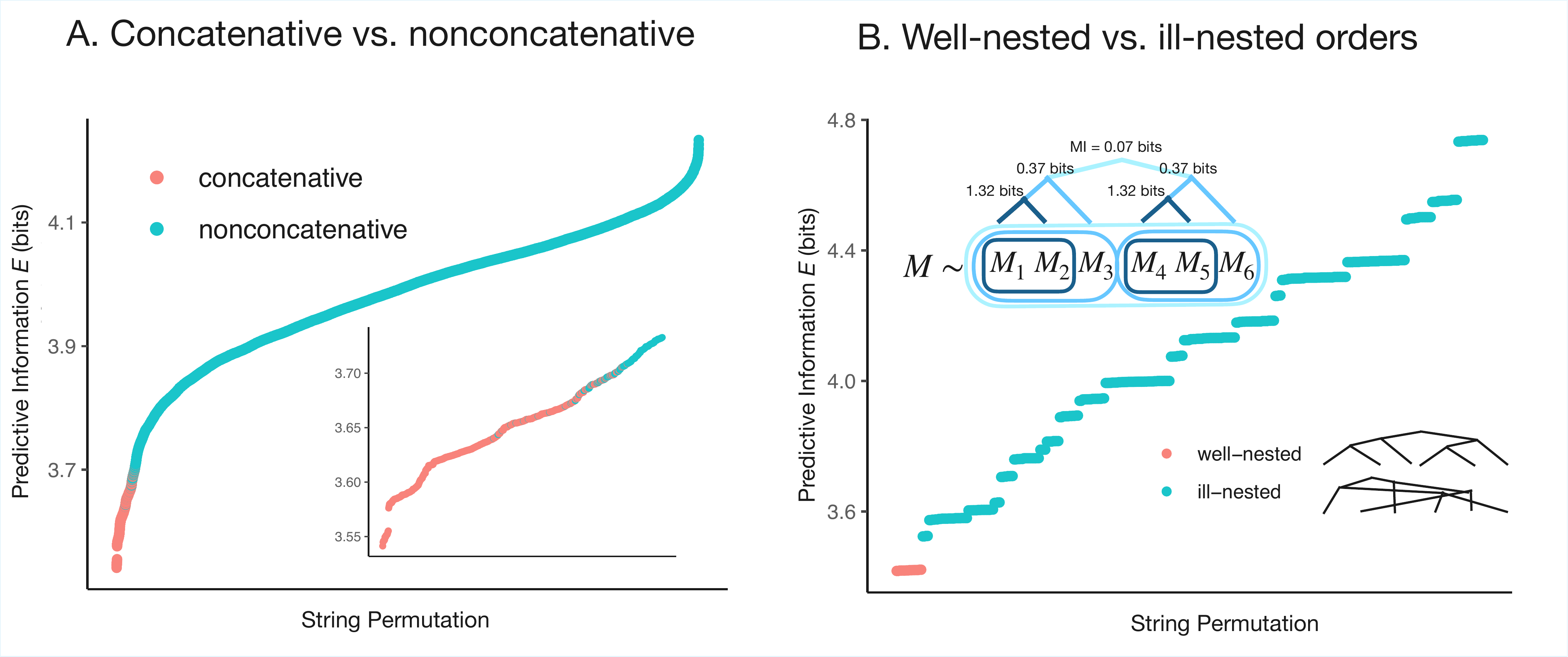}
\caption{Simulations of codes with different orders or elements. \textbf{A.} Predictive information of all string permutations of a systematic language for a Zipfian source. Permutations that combine components by concatenation, marked in red, achieve the lowest predictive information. The inset zooms in on the 2000 permutations with the lowest predictive information.
\textbf{B.} A hierarchically-structured source distribution (see text) and predictive information of all permutations of a systematic language for this source. A language is well-nested when all groups of letters corresponding to groupings in the inset tree figure are contiguous.}
\label{fig:order-sims}
\end{figure}

\paragraph{Locality} Next, we show that minimization of predictive information yields languages where features of meaning correspond to localized parts of strings, corresponding to words. We consider a Zipfian distribution over 100 meanings, and a language $L$ in which forms consist of two length-4 `words'. We then consider scrambled languages formed by applying permutations to the string output of $L$. For example, if the original language expresses a meaning with two words like $L(m_1 \times m_2) = \texttt{aaaa} \cdot \texttt{bbbb}$, a possible scrambled language would have $L^\prime(m_1 \times m_2) = \texttt{baaabbab}$. These scrambled languages instantiate possible string combination functions other than concatenation.

Calculating predictive information for all possible scrambled languages, we find that the languages in which the `words' remain contiguous have the lowest predictive information, as shown in Fig.~\ref{fig:order-sims}A. This happens because the coding procedure above creates correlations among letters within a word. When these correlated letters are separated from each other---such as when letters from another word intervene---then predictive information increases. Interestingly, not every concatenative language is better than every nonconcatenative one. This corresponds to the reality of natural language, in which limited nonconcatenative and nonlocal morphophonological processes do exist, for example, in Semitic nonconcatenative morphology\cite{mccarthy1981prosodic}.

\paragraph{Hierarchical structure} Natural language sentences typically have well-nested hierarchical syntactic structures, of the kind generated by a context-free grammar\citep{chomsky1957syntactic}: for example the sentence \emph{[[the big dog] chased [a small cat]]} has two noun phrases, indicated by brackets, which are contiguous and nested within the sentence. Minimization of predictive information creates these well-nested word orders, with phrases corresponding to groups of words that are more or less strongly correlated\citep{mansfield2023emergence}. We demonstrate this effect using a source distribution defined over six random variables $M_1, \dots, M_6$ with a covariance structure shown in the inset of Fig.~\ref{fig:order-sims}B: each of the variable pairs $(M_1, M_2)$ and $(M_4, M_5)$ are highly internally correlated; these pairs are weakly correlated with $M_3$ and $M_6$ respectively; and both groups of variables are very weakly correlated with each other. As above, we consider all possible permutations of a systematic code for these source variables. The codes that minimize predictive information are those which are well-nested with respect to the correlation structure of the source, keeping the letters corresponding to all groups of correlated features contiguous. Further simulation results involving context-free languages are found in SI~Section~G. For a mathematical analysis of predictive information in local and random orders for structured sources, see SI~Section~A.


\subsection*{Cross-Linguistic Empirical Results}

Here we present cross-linguistic empirical evidence that the systematic structure of language has the effect of reducing predictive information at the levels of phonotactics, morphology, syntax, and semantics, compared against systems that lack natural local systematicity.

\begin{figure}
\centering
\includegraphics[width=.9\textwidth]{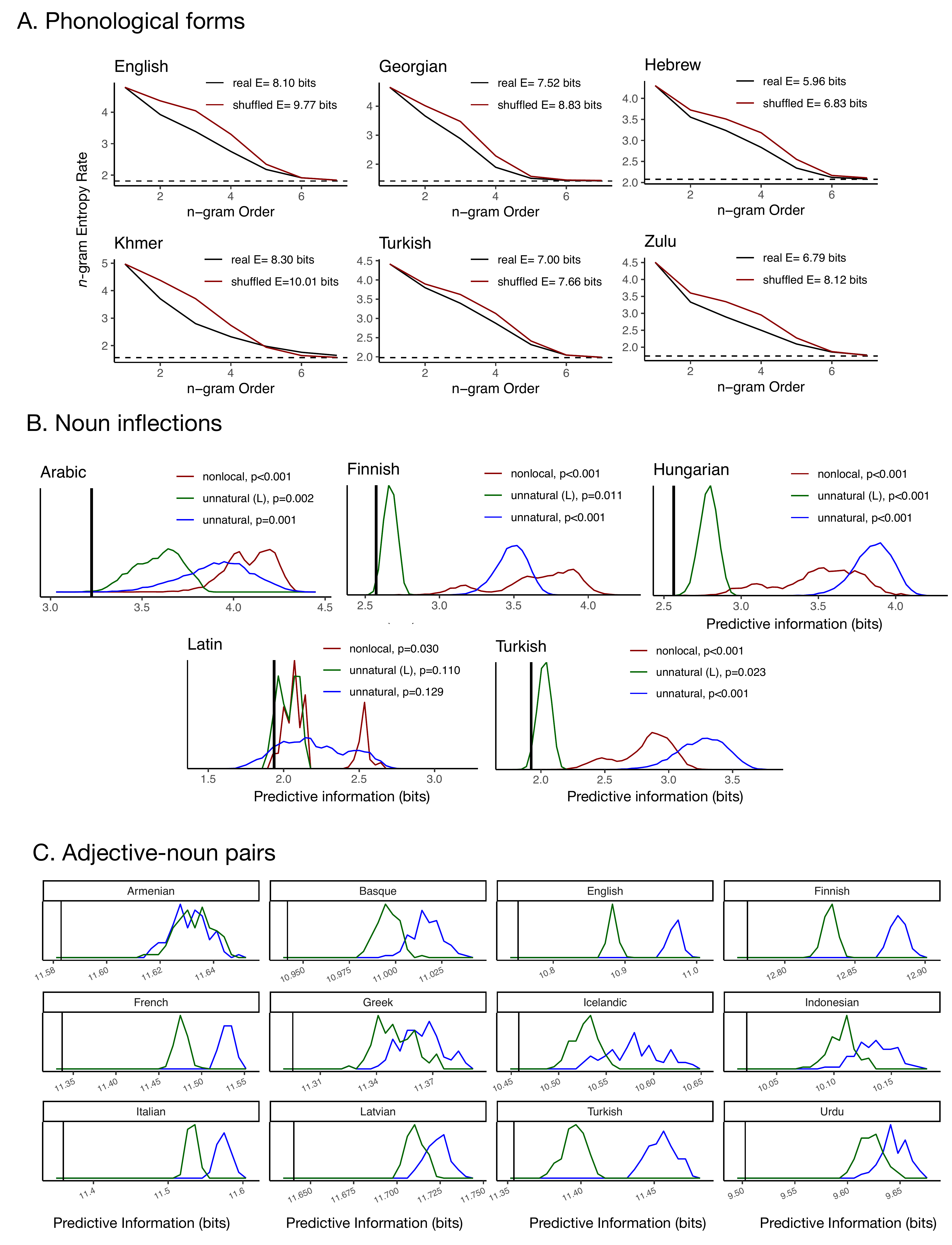}
\caption{Evidence that natural languages are configured in a way that reduces predictive information, in phonotactics, morphology, and syntax.
\textbf{A.} Predictive information calculation for phonological forms in selected languages, comparing the attested forms against forms that have been deterministically shuffled while preserving manner of articulation. 
\textbf{B.} Letter-level predictive information of noun morphology (black vertical line), compared against predictive information values for four random baselines (densities; see text). $p$ values indicate the proportion of baseline samples with lower predictive information than the attested forms.
\textbf{C.} Letter-level predictive information of adjective--noun pairs from 12 languages, compared with baselines. (Nonlocal baselines always generate much higher predictive information than the attested forms, and are not shown.)
}
\label{fig:corpus_figures}
\end{figure}

\paragraph{Phonotactics} Languages have restrictions on what sequences of sounds may occur within words: for example, \emph{blick} seems like a possible English word, whereas \emph{bnick} does not, even though it is pronounceable in other languages\citep{chomsky1968sound}. These systems of restrictions are called phonotactics. Here we show that actual phonotactic systems of human languages, which involve primarily local constraints on what sounds may co-occur, result in lower predictive information compared to counterfactual phonotactic systems. We compare phonemically-transcribed wordforms in vocabulary lists of 61 languages against counterfactual alternatives generated by deterministically scrambling phonemes within a word while preserving manner of articulation. This ensures that the resulting counterfactual forms are roughly possible to articulate. For example, an English word \emph{fasted} might be scrambled to form \emph{sefdat}. Calculating predictive information, we find that the real vocabulary lists have lower predictive information than the counterfactual variants in all languages tested. Results for 6 languages with diverse sound systems are shown in Fig.~\ref{fig:corpus_figures}A. Results for the remaining 55 languages are in SI~Appendix~C. 

\begin{figure}
\centering
\includegraphics[width=\textwidth]{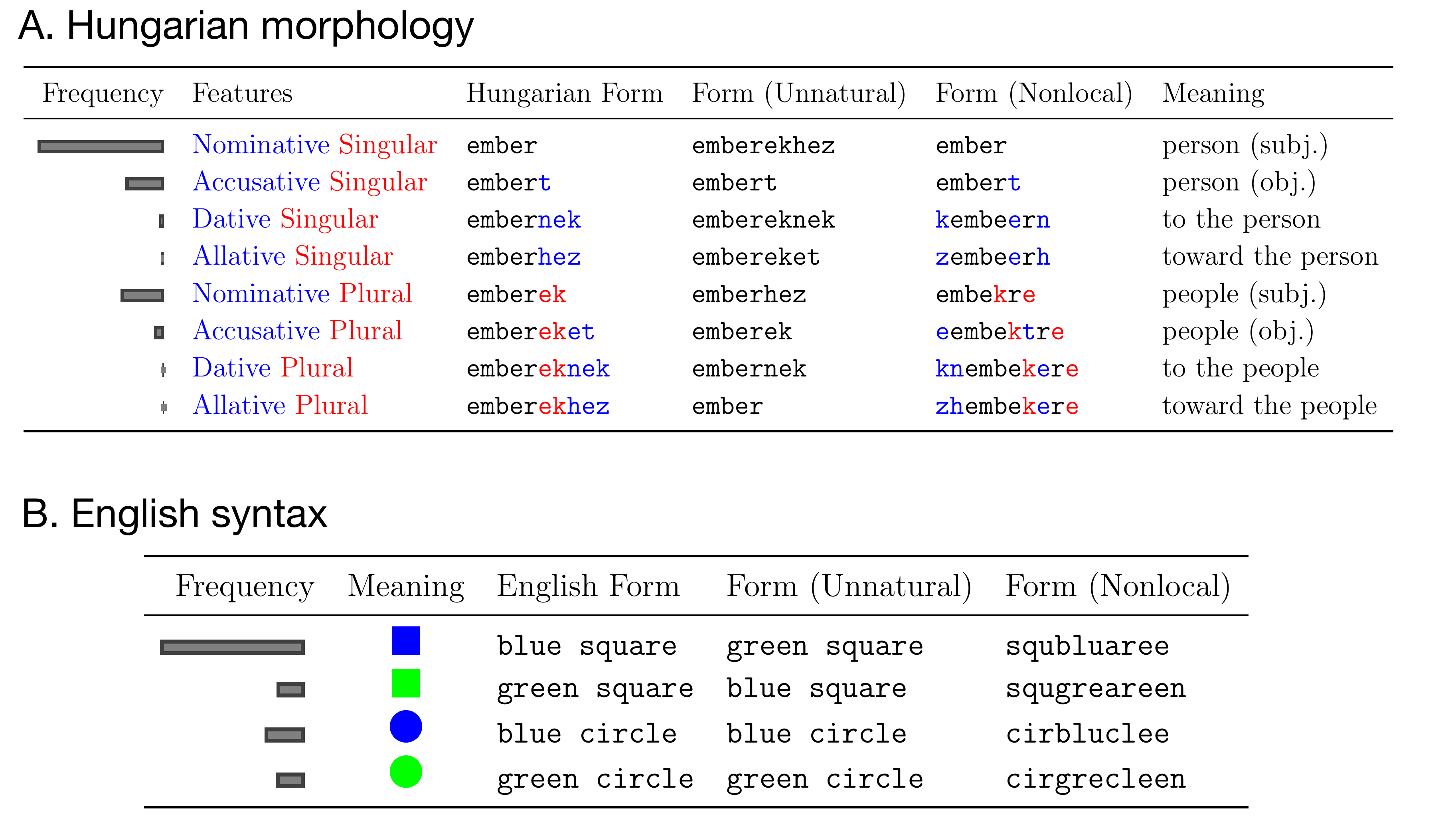}
\caption{Examples of systematic morphology and syntax, and baselines used in experiments.  
\textbf{A.} Forms of the Hungarian noun \texttt{ember} meaning `person', along with examples of the unnatural and nonlocal baseline used in Fig.~\ref{fig:corpus_figures}B. 231 additional forms not shown. `Frequency' column illustrates the joint frequency of grammatical features in the Hungarian Szeged UD corpus\citep{vincze2010hungarian,nivre2015universal}.
\textbf{B.} English forms for the given meanings, along with frequencies from the English Common Crawl web corpus\citep{buck2014ngram}. Example unnatural and nonlocal baseline forms are shown.}
\label{fig:synmorph-tables}
\end{figure}

\paragraph{Morphology} Words change form to express grammatical features in a way that is often systematic. For example, the forms of the Hungarian noun shown in Fig.~\ref{fig:synmorph-tables}A are locally systematic with respect to case and number features. In Fig.~\ref{fig:corpus_figures}B, we show that the local systematic structure of affixes for case, number, possession, and definiteness in 5 languages has the effect of reducing predictive information when comparing against baselines that disrupt this structure. We estimate predictive information of these morphological affixes across five languages, with source distributions proportional to empirical corpus counts of the joint frequencies of grammatical features. We compare the predictive information of the attested forms against three alternatives: (1) a nonlocal baseline generated by applying a deterministic permutation function to each form, (2) an unnatural baseline generated by permuting the assignment of forms to meanings (features); and (3) a more controlled unnatural baseline which permutes the form--meaning mapping while preserving form length. The unnatural baselines preserve the phonotactics of the original forms; only the form--meaning relationship is changed.

Across the languages, we find that the attested forms have lower predictive information than the majority of baselines. The weakest effect is in Latin, which also has the most fusional and least systematic morphology\citep{rathi2021information}. Note that Arabic nouns often show nonconcatenative morphology in the form of so-called `broken' plurals: for example, the plural of the loanword \emph{film} meaning `film' is \emph{'aflām}. This pattern is represented in the forms used to generate Fig.~\ref{fig:corpus_figures}B, and yet Arabic noun forms still have lower predictive information than the majority of baseline samples. This suggests that the limited form of nonconcatenative morphology present in Arabic is still consistent with the idea that languages are configured in a way that keeps predictive information low.

\paragraph{Syntax}
Phrases such as \texttt{blue square} have natural local systematicity, as shown in Fig.~\ref{fig:synmorph-tables}B. We compare real adjective--noun combinations in corpora of 12 languages against unnatural and nonlocal baselines generated the same way as in the morphology study: permuting the letters within a form to disrupt locality, or permuting the assignment of forms to meanings to disrupt naturalness. We estimate the probability of a meaning as proportional to the frequency of the corresponding adjective--noun pair. Results are shown in Fig.~\ref{fig:corpus_figures}C. The real adjective--noun pairs have lower predictive information than a large majority of baselines across all languages tested. 

\begin{figure}
\centering
\includegraphics[width=\textwidth]{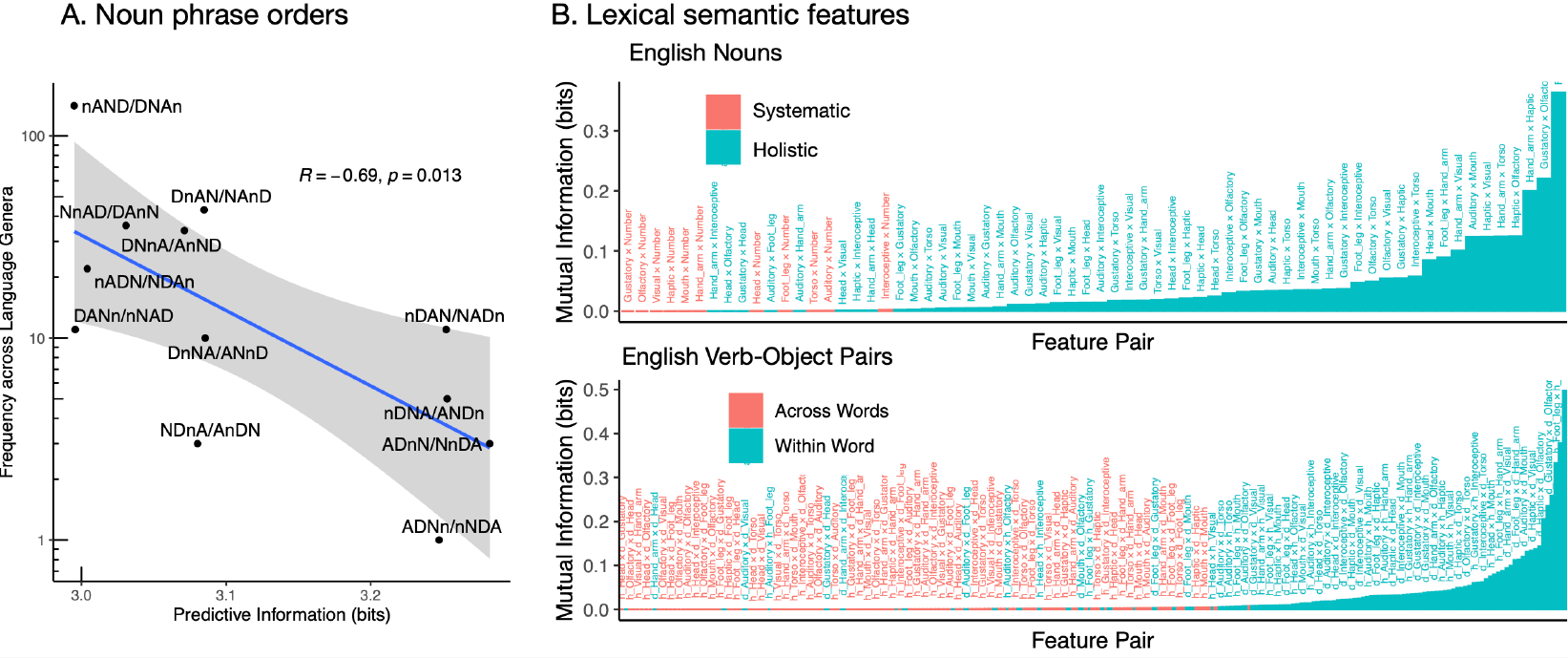}
\caption{Evidence that word order and lexical semantics are configured in ways that reduce predictive information.
\textbf{A.} Typological frequency of noun phrase orders (number of unrelated language genera showing the given order\citep{dryer2018order}) as a function of predictive information. More frequent orders have lower predictive information. Blue line shows a linear regression predicting log typological frequency from predictive information. Error bars indicate a 95\% confidence interval of the slope of this regression. The negative correlation is significant with Pearson's $R=-0.69$ and $p=0.013$. 
\textbf{B.} Top: Pairwise mutual information of semantic features from the Lancaster Sensorimotor Norms\citep{lynott2020lancaster} in addition to a number feature, as indicated by plural morphology. The number feature is expressed systematically; all others are holistic. Bottom: Pairwise mutual information values for Lancaster Sensorimotor Norm features across and within words, for pairs of verbs and their objects.}
\label{fig:corpus_figures2}
\end{figure}

\paragraph{Word Order}
In an English noun phrase such as \texttt{the three cute cats}, the elements Determiner (D, \texttt{the}), Numeral (N, \texttt{three}), Adjective (A, \texttt{cute}), and Noun (n, \texttt{cats}) are combined in the order D--N--A--n. This order varies across languages---for example, Spanish has D--N--n--A (\texttt{los tres gatos lindos})---but certain orders are more common than others\citep{dryer2018order}. We aim to explain the cross-linguistic distribution of these orders through reduction of predictive information, which drives words that are statistically predictive of each other to be close to each other, an intuition shared with existing models of adjective order\citep{futrell2019information,culbertson2020world,mansfield2023emergence}. To do so, we estimate source probabilities for noun phrases (consisting of single head lemmas for a noun along with an optional adjective, numeral, and determiner) based on corpus frequencies. We then calculate predictive information at the word level (treating words as single atomic symbols) for all possible permutations of D--N--A--n. Predictive information is symmetric with respect to time reversal, so we cannot distinguish orders such as D--N--A--n from n--A--N--D, etc. As shown in Fig.~\ref{fig:corpus_figures2}A, the orders with lower predictive information are also the orders that are more frequent cross-linguistically. A number of alternative source distributions also yield this downward correlation, as shown in SI~Appendix~D. 

\paragraph{Lexical Semantics}
Considering a word such as \texttt{cats}, all the semantic features of a cat (furriness, mammalianness, etc.) are expressed holistically in the morpheme \texttt{cat}, while the feature of numerosity is separated into the plural marker \texttt{--s}. Plural marking like this is common across languages\citep{corbett2000number}. From reduction of predictive information, we expect relatively uncorrelated components of meaning to be expressed systematically, and relatively correlated components to be expressed together holistically. Thus, we hold that numerosity is selected to be expressed systematically in a separate morpheme because it is relatively independent of the other features of nouns, which are in turn highly correlated with each other. Our theory thus derives the intuition that natural categories arise from the correlational structure of experience\citep{rosch1978principles}.

We validate this prediction in a study of semantic features in English, using the Lancaster Sensorimotor Norms\citep{lynott2020lancaster} to provide semantic features for English words and using the English Universal Dependencies (UD) corpus to provide a frequency distribution over words. The Lancaster Sensorimotor Norms provide human ratings for words based on sensorimotor dimensions, such as whether they involve the head or arms. As shown in Fig.~\ref{fig:corpus_figures2}B (top), we find that the semantic norm features are highly correlated with each other, and relatively uncorrelated with numerosity, as predicted by the theory.

For the same reason, the theory also predicts that semantic features should be more correlated within words than across words. In Fig.~\ref{fig:corpus_figures2}B (bottom), we show within-word and cross-word correlations of the semantic norm features for pairs of verbs and their objects taken from the English UD corpus. As predicted, the across-word correlations are weaker. Correlations based on features drawn from other semantic norms are presented in SI~Appendix~E.

\section*{Discussion}

Our results underscore the fundamental roles of prediction and memory in human cognition and provide a link between the algebraic structure of human language and information-theoretic concepts used in machine learning and neuroscience. Our work joins the growing body of information-theoretic models of human language based on resource-rational efficiency\citep{ferrericancho2003least,jaeger2011language,kemp2012kinship,zaslavsky2018efficient,gibson2019efficiency,levshina2022communicative}.

\paragraph{Language models} Large language models are based on neural networks trained to predict the next token of text given previous tokens. Our results suggest that language is structured in a way that makes this next-token prediction relatively easy, by minimizing the amount of information that needs to be extracted from the previous tokens to predict the following tokens. Although it has been claimed that large language models have little to tell us about the structure of human language---because their architectures do not reflect formal properties of grammars and because they can putatively learn unnatural languages as well as natural ones\citep{mitchell2020priorless,chomsky2023nyt,moro2023impossible}---our results suggest that these models have succeeded so well precisely because natural language is structured in a way that makes their prediction task relatively simple. Indeed, neural sequence architectures struggle to learn languages that lack information locality\citep{kallini2024mission,someya2025information}.

\paragraph{Machine learning}
Our results establish a connection between the structure of human language and ideas from machine learning. In particular, minimization of mutual information (a technique known as Independent Components Analysis, ICA\cite{ans1985architectures,bell1995information}) is widely deployed to create representations that are `disentangled' or compositional\cite{bengio2013representation}, and to detect object boundaries in images, under the assumption that pixels belonging to the same object exhibit higher statistical dependence than pixels belonging to different objects\citep{isola2014crisp}. (Although general nonlinear ICA with real-valued outputs does not yield unique solutions\citep{hyvarinen1999nonlinear}, we have found above that minimization of predictive information does find useful structure in our setting, with discrete string-valued outputs and a deterministic function mapping meaning to form.) We propose that human language follows a similar principle: it reduces predictive information, which amounts to performing a generalized sequential ICA on the source distribution on meanings, factoring it into groups of relatively independent components that are expressed systematically as words and phrases, with more statistical dependence within these units than across them. This provides an explanation for why ICA-like objectives yield representations that are intuitively disentangled, compositional, or interpretable: they yield the same kinds of concepts that we find encoded in natural language. 

\paragraph{Neuroscience}
Similarly, neural codes have been characterized as maximizing information throughput subject to information-theoretic and physiological constraints\cite{linsker1988self,stone2018principles}, including explicit constraints on predictive information\cite{bialek2006efficient,palmer2015predictive}. These models predict that, in many cases, neural codes are \emph{decorrelated}: distinct neural populations encode statistically independent components of sensory input\cite{barlow1989unsupervised}. 
Our results suggest that language operates on similar principles: it expresses meanings in a way that is temporally decorrelated.
This view is compatible with neuroscientific evidence on language processing: minimization of predictive information (while holding overall predictability constant) equates to maximization of local predictability of the linguistic signal, a driver of the neural response to language\cite{schrimpf2021neural,heilbron2022hierarchy}.

\paragraph{Information theory and language}

Previous work\citep{hahn2021modeling} derived locality in natural language from a related information-theoretic concept, the memory--surprisal tradeoff or predictive information bottleneck curve, which describes the best achievable sequential predictability as a function of memory usage\citep{still2014information}. The current theory is a simplification that looks at only one part of the curve: predictive information is the minimal memory at which sequential predictability is maximized. A more complete information-theoretic view of language may have to consider the whole curve.

We join existing work attempting to explain linguistic structure on the basis of information-theoretic analysis of language as a stochastic process, for example,  the study of lexical scaling laws as a function of redundancy and nonergodicity in text\citep{debowski2011vocabulary}. Other work on predictive information in language has focused on the long-range scaling of the $n$-gram entropy in connected texts, with results seeming to imply that the predictive information diverges\cite{debowski2011excess,debowski2015relaxed}. In contrast, we have focused on only single utterances, effectively considering only relatively short-range predictive information.

\paragraph{Cognitive status of predictive information} 
Predictive information is a fundamental measure of complexity, which may manifest explicitly or implicitly in various ways in the actual mechanisms of language production, comprehension, and learning. For example, in a recent model of online language comprehension\citep{hahn2022resource}, comprehenders predict upcoming words on the basis of memory representations that are constrained to store only a small number of words. The fundamental limits of predictive information apply implicitly in this model because comprehenders' predictions cannot be more accurate than if they stored an equivalent amount of predictive information. As another example, a model of language production based on short stored chunks\citep{mansfield2023emergence} would effectively produce language with low predictive information, because these chunks would be relatively independent of each other, while predictive relationships inside the stored chunks would be preserved. Predictive information has also been linked to difficulty of learning: processes containing more predictive information require more parameters and data to be learned\citep{bialek2001predictability}, and any learner with limited ability to learn long-term dependencies will have an effective inductive bias toward languages with low predictive information. Predictive information is not meant as a complete model of the constraints on language, which would certainly involve factors beyond predictive information as well as separate, potentially competing pressures from comprehension and production\citep{dell2003neighbors}.

Relatedly, while we have shown that natural language is configured in a way that keeps predictive information low, we have not speculated on \emph{how} languages come to be configured in this way, in terms of language evolution and change. We believe there are multiple pathways for this to happen. For example, efficiency pressures in individual interactions could give rise to overall efficient conventions\citep{hawkins2023partners}, or memory limits in learning\citep{newport1990maturational,cochran1999too} could cause learners to form low-predictive-information generalizations from their input. Identifying the causal mechanisms that control predictive information is a critical topic for future work.

\paragraph{Linguistics}
Our theory of linguistic systematicity is independent of theoretical assumptions about mental representations of grammars, linguistic forms, or the meanings expressed in language. Predictive information is a function only of the probability distribution on forms, seen as one-dimensional sequences of symbols unfolding in time. This independence from representational assumptions is an advantage, because there is as yet no consensus about the basic nature of the mental representations underlying human language\cite{jackendoff2007linguistics,goldberg2009constructions}. 

Our results reflect and formalize a widespread intuition about human language, first formulated as Behaghel's Law\cite{behaghel1932deutsche}: ``that which is mentally closely related is also placed close together''. For example, words are contiguous units and the order of morphemes within them is determined by a principle of relevance\cite{bybee1985morphology,givon1991isomorphism}, and important aspects of word order across languages have been explained in terms of dependency locality, a principle that syntactically linked words are close\cite{hawkins2004efficiency,liu2017dependency,temperley2018minimizing,futrell2020dependency}. 

A constraint on predictive information predicts information locality: elements of a linguistic form should be close to each other when they \emph{predict} each other\cite{futrell2019information}.
We propose that information locality subsumes existing intuitive locality ideas.
Thus, since words have a high level of statistical inter-predictability among their parts\cite{mansfield2021word}, they are mostly contiguous, and as a residual effect of this binding force, related words are also close together.
Furthermore, we have found that the same formal principle predicts the existence of linguistic systematicity and the way that languages divide the world into natural kinds\cite{rosch1978principles,barrett2007dynamic}.

\paragraph{Limitations} Much work is required to push our hypothesis to its limit. We have assumed throughout that languages are one-to-one mappings between form and meaning; the behavior of ambiguous or nondeterministic codes, where ambiguity might trade off with predictive information, may yield additional insight. Furthermore, we have examined predictive information only within isolated utterances. It remains to be seen whether reduction of predictive information, applied at the level of many connected utterances, would be able to explain aspects of discourse structure such as the hierarchical organization of topics and topic--focus structure\citep{chafe1976givenness}.

There remain several aspects of human language that remain unexplained under our theory. One known limitation is that predictive information is symmetric with respect to time reversal, so (at least when applied at the utterance level) it cannot explain time-asymmetric properties of language such as the pattern of `accessible' (frequent, animate, definite, and given) words appearing earlier within utterances than inaccessible ones\citep{bock1982toward,bresnan2007predicting}. There is also the fact that non-local and non-concatenative structures do exist in language, for example, long-term coreference relationships among discourse entities, and long-distance filler--gap dependencies, which would seem to contravene the idea that predictive information is constrained. An important area for future research will be to determine what effect these structures really have on predictive information, and what other constraints on language might explain them.

\section*{Methods}

\paragraph{Constructing a stochastic process from a language} We define a language as a mapping from a set of meanings to a set of strings, $L : \mathcal{M} \rightarrow \Sigma^*$. In order to define predictive information \emph{of a language}, we need a way to derive a stationary stochastic process generated by that language. We use the following mathematical construction which generates an infinite stream of symbols: (1) meanings $m \sim p_M$ are sampled iid from the source distribution $p_M$, (2) each meaning is translated into a string as $s = L(m)$, (3) the strings $s$ are concatenated end-to-end in both directions with a delimiter $\# \notin \Sigma$ between them.
Finally, a string is chosen with probability reweighted by its length, and a time index $t$ (relative to the closest delimiter to the left) is selected uniformly at random within this form.

This construction has the effect of zeroing out any mutual information between symbols with the delimiter between them. Thus, when we compute $n$-gram statistics, we can treat each form as having infinite padding symbols to the left and right. This is the standard method for collecting $n$-gram statistics in natural language processing\cite{chen1999empirical}. 

\paragraph{Three-feature source simulation}
For Figs.~\ref{fig:coin-sims}B and~\ref{fig:coin-sims}C, the source distribution is
\begin{equation}
\label{eq:indep-bern-source}
M \sim \text{Bernoulli}\left(\frac{2}{3}\right) \times \text{Bernoulli}\left(\frac{2}{3} + \varepsilon\right) \times \text{Bernoulli}\left(\frac{2}{3} + 2\varepsilon\right),
\end{equation}
with $\varepsilon=.05$. 

For Fig.~\ref{fig:coin-sims}E, we need to generate distributions of the form $p(M) = p(M_1) \times p(M_2, M_3)$ while varying the mutual information $\operatorname{I}[M_2 : M_3]$. We start with the source from Eq.~\ref{eq:indep-bern-source} (whose components are here denoted $p_{\text{indep}}$) and mix it with a source that creates a correlation between $M_2$ and $M_3$:
\begin{equation}
p_\alpha(M=ijk) = p_{\text{indep}}(M_1=i) \times \left[\left(1-\alpha\right)\left(p_{\text{indep}}(M_2=j) \times p_{\text{indep}}(M_3=k)\right) + \frac{\alpha}{2}\delta_{jk}\right],
\end{equation}
with $\delta_{jk} = 1$ if $j=k$ and $0$ otherwise. The mixture weight $\alpha$ controls the level of mutual information, ranging from $0$ at $\alpha=0$ to at most $1$ bit at $\alpha=1$. A more comprehensive study of the relationship between feature correlation, systematicity, and predictive information is given in SI~Appendix~B, which examines systematic and holistic codes for a comprehensive grid of possible distributions on the simplex over 4 outcomes.


\paragraph{Locality simulation}
For the simulation shown in Fig.~\ref{fig:order-sims}A, we consider a source over 100 objects labeled $\{m^{00}, m^{01},\dots, m^{99}\}$, following a Zipfian distribution $p(M=m^i) \propto \left(i+1\right)^{-1}$. We consider a language based on a decomposition of the meanings based on the digits of their index, with for example $m^{89}$ decomposing into features as $m_1^8 \times m_2^9$. Each utterance decomposes into two `words' as $L(m_1 \times m_2) = L(m_1) \cdot L(m_2)$, where the word for each feature $m^k$ is a random string in $\{\texttt{0},\texttt{1}\}^4$, maintaining a one-to-one mapping between features $m^k$ and words.

\paragraph{Hierarchy simulation}
For the simulation shown in Fig.~\ref{fig:order-sims}B, we consider a source $M$ over $5^6 = 15625$ meanings which may be expressed in terms of six random variables $\left\langle M_1, M_2, M_3, M_4, M_5, M_6\right\rangle$ each over 5 outcomes, with a probability distribution as follows:
\begin{align}
p(M) &= \alpha q(M_1, M_2, M_3, M_4, M_5, M_6)+ \left(1 - \alpha\right)\bigg( \\
\nonumber
&\left[\beta q(M_1, M_2, M_3) + \left(1-\beta\right)\left[\gamma q(M_1, M_2) + \left(1-\gamma\right)q(M_1)q(M_2) \right]    q(M_3) \right] \\
\nonumber
&\times \left[\beta q(M_4, M_5, M_6) + \left(1-\beta\right)\left[\gamma q(M_4, M_5) + \left(1-\gamma\right)q(M_4)q(M_5) \right]q(M_6)    \right]\bigg),
\end{align}
where $\alpha=.01$, $\beta=0.20$, $\gamma=0.99$ are coupling constants, and each $q(\cdot)$ is a Zipfian distribution as above. The coupling constants control the strengths of the correlations shown in Fig.~\ref{fig:order-sims}B.

\paragraph{Phonotactics} We assume a uniform distribution over forms found in WOLEX. SI~Section~F shows results for four languages using corpus-based word frequency estimates to form the source distribution, with similar results.

\paragraph{Morphology} We estimate the source distribution on grammatical features (number, case, possessor, and definiteness) using the feature annotations from UD corpora, summing over all nouns, with add-$\sfrac{1}{2}$ smoothing. The dependency treebanks are drawn from UD v2.8\cite{nivre2015universal}: for Arabic, NYUAD Arabic UD Treebank; for Finnish, Turku Dependency Treebank; for Turkish, Turkish Penn Treebank; for Latin, Index Thomisticus Treebank; for Hungarian, Szeged Dependency Treebank. Forms are represented with a dummy symbol \texttt{X} standing for the stem, and then orthographic forms for suffixes, for example $\texttt{Xoknak}$ for the Hungarian dative plural. For Hungarian, Finnish, and Turkish, we use the forms corresponding to back unrounded vowel harmony. For Latin, we use first-declension forms. For Arabic, we use regular masculine triptote forms with a broken plural; to do so, we represent the root using three dummy symbols, and the plural using a common `broken' form\citep{thackston1994koranic}, with for example \texttt{XaYZun} for the nominative indefinite singular and \texttt{'aXYāZun} for the nominative indefinite plural. Results using an alternate broken plural form \texttt{XiYāZun} are nearly identical.

\paragraph{Adjective--noun pairs} From UD corpora we extract adjective--noun pairs, defined as a head wordform with part-of-speech \texttt{NOUN} modified by an adjacent dependent wordform with relation \texttt{amod} and part-of-speech \texttt{ADJ}. The forms over which predictive information is computed consist of the pair of adjective and noun from the corpus, in their original order, in original orthographic form with a whitespace between them. The source distribution is directly proportional to the frequencies of the forms.

\paragraph{Noun phrase order} The source distribution on noun phrases is estimated from the empirical frequency of noun phrases in the German GSD UD corpus, which has the largest number of such noun phrases among the UD corpora. To estimate this source, we define a noun phrase as a head lemma of part-of-speech \texttt{NOUN} along with the head lemmas for all dependents of type \texttt{amod} (with part-of-speech \texttt{ADJ}), \texttt{nummod} (with part-of-speech \texttt{NUM}), and \texttt{det} (with part-of-speech \texttt{DET}). We extract these noun phrase forms from the corpus. When a noun phrase has multiple adjectives, one of the adjectives is chosen randomly and the others are discarded. The result is counts of noun phrases of the form below: \\ 
\begin{center}
\begin{tabular}{llllr}
\toprule
Det & Num & Adj & Noun & Count\\
\midrule
die & --- & --- & Hand & 234 \\
ein & --- & alt & Kind & 4 \\
--- & drei & --- & Buch & 2 \\
ein & --- & einzigartig & Parf\"umeur & 1\\
\dots & \dots & \dots & \dots & \dots \\
\bottomrule
\end{tabular}
\end{center}

The source distribution is directly proportional to these counts. We then compute predictive information at the word level over the attested noun phrases for all possible permutations of determiner, numeral, adjective, and noun. Typological frequencies are as given by Dryer (2018)\citep{dryer2018order}.

\paragraph{Semantic features} We binarize the Lancaster Sensorimotor Norms\cite{lynott2020lancaster} by recoding each norm as 1 if it exceeds the mean value for that feature across all words, and 0 otherwise. Word frequencies are calculated by maximum likelihood based on lemma frequencies in the concatenation of the English GUM\cite{zeldes2017gum}, GUMReddit\cite{behzad2020crossgenre}, and EWT\cite{silveira2014gold} corpora from UD 2.8. The `Number' feature is calculated based on the value of the \texttt{Number} feature in the UD annotations. Verb--object pairs were identified as a head wordform with part-of-speech \texttt{VERB} with a dependent wordform of relation \texttt{obj} and part-of-speech \texttt{NOUN}.

\section*{Data availability} Unique data required to reproduce our results is available at \url{http://github.com/Futrell/infolocality}. Corpus count data is drawn from Universal Dependencies v2.8, available at \url{https://lindat.mff.cuni.cz/repository/xmlui/handle/11234/1-3683}. The Lancaster Sensorimotor Norms are available at \url{https://osf.io/7emr6/}. Wordform data from the WOLEX database\citep{graff2012communicative} is not publicly available, but a subset can be made available upon request to the authors.

\section*{Code availability} Code to reproduce our results is available at \url{http://github.com/Futrell/infolocality}.

\section*{Acknowledgments} We thank 
Steve Piantadosi, Neil Rathi, Greg Scontras, Kyle Mahowald, Noga Zaslavsky, Tiago Pimentel, Robert Hawkins, Nathaniel Imel, Ruichen Sun, Zygmunt Pizlo, Brian Skyrms, Jeff Barrett, Jacob Andreas, Matilde Marcolli, Juan Pablo Vigneaux Ariztia, {\L}ukasz D\k{e}bowski, Andrea Nini, and audiences at NeurIPS InfoCog 2023, the UCI Center for Theoretical Behavioral Sciences, EvoLang 2024, TedLab, the Society for Computation in Linguistics 2024, the Quantitative Cognitive Linguistics Network, and the CalTech Seminar on Information and Geometry for discussion. The authors received no specific funding for this work.
 
\section*{Author contributions} RF designed and ran studies in the main text. RF and MH performed mathematical analyses, designed and ran studies in the Supplementary Information, and wrote the manuscript.

\section*{Competing interests} The authors declare no competing interests.

\end{document}


\title{Supplemental Information: Linguistic Structure from a Bottleneck on Sequential Information Processing}








\appendix

\tableofcontents

\section{Basic formal results}
\label{app:formal-results}

In the Main Text, based on numerical simulations, we claimed that codes that minimize predictive information tend to (1) factorize their source distribution into approximately independent components, and (2) express these components systematically in local parts of strings. Here we provide some elementary theorems about codes that minimize predictive information, which illustrate these generalizations. Although a full mathematical analysis of such codes is beyond the scope of the current work, the results below serve to establish their general behavior.

\subsection{Forms of predictive information}

In the Main Text, we claimed that the predictive information of a stochastic process can be thought of in terms of successive approximations to the entropy rate based on $n$-gram models with successively larger context size $n$. This has previously been shown by \citet[][Prop. 8]{crutchfield2003regularities}, among others. Here we provide the same result by means of a different and more direct proof.

Consider a stationary stochastic process generating symbols labelled $\dots, X_{t-1}, X_t, X_{t+1}, \dots$ extending into the infinite past and future. Predictive information is defined as the limit of the mutual information of large blocks of $M$ symbols before and $N$ symbols after an arbitrary time index $t$:
\begin{equation}
\label{eq:ee-mi}
E = \lim_{N \rightarrow \infty} \lim_{M \rightarrow \infty} \mathrm{I}[X_{t-M:t} : X_{t:t+N}],
\end{equation}
where $X_{a:b} = X_a, \dots, X_{b-1}$ represents a block of symbols $X$ indexed by an exclusive range. We write Eq.~\ref{eq:ee-mi} in shorthand as the mutual information between the infinite past $X_{<t}$ and infinite future $X_{\ge t}$ of the process,
\begin{equation}
E = \mathrm{I}[X_{<t} : X_{\ge t}].
\end{equation}

Now we can state the theorem relating predictive information to entropy rates derived from $n$-gram models.

\begin{theorem} 
The predictive information $E$ can be written as
\begin{equation}
\label{eq:ee-rates}
E = \lim_{N \rightarrow \infty} \sum_{n=1}^N \left(h_n - h\right),
\end{equation}
where $h_n$ is the $n$-gram entropy rate
\begin{align}
h_n &= \mathrm{H}[X_t \mid X_{t-n+1:t}],
\end{align}
and $h$ is the asymptotic entropy rate
\begin{equation}
h = \lim_{n \rightarrow \infty} h_n.
\end{equation}
\end{theorem}
\begin{proof}
Invoking stationarity, we set $t=1$ without loss of generality. Using the chain rule for mutual information, we rewrite the predictive information as a sum of conditional mutual informations:
\begin{equation}
\label{eq:chain-rule}
E = \lim_{N \rightarrow \infty} \lim_{M \rightarrow \infty} \sum_{n=1}^N \mathrm{I}[X_{1-M:1} : X_n \mid X_{1:n}].
\end{equation}
Now we break each mutual information term into a difference of conditional entropies:
\begin{equation}
E = \lim_{N \rightarrow \infty} \lim_{M \rightarrow \infty} \sum_{n=1}^N  \left( \mathrm{H}[X_{n} \mid X_{1:n}] - \mathrm{H}[X_{n} \mid X_{1-M:n}] \right).
\end{equation}
Because conditioning reduces entropy, the terms $\mathrm{H}[X_{n} \mid X_{1-M:n}]$ (which are finite) converge monotonically downward in $M$, so we may swap the sum and the limit on $M$. Then, invoking stationarity again, we notice that the resulting two terms are the $n$-gram entropy rate and the asymptotic entropy rate:
\begin{align}
E &= \lim_{N \rightarrow \infty} \sum_{n=1}^N  \left( \underbrace{\mathrm{H}[X_{n} \mid X_{1:n}]}_{\text{$n$-gram entropy rate}} - \underbrace{\lim_{M \rightarrow \infty} \mathrm{H}[X_{n} \mid X_{1-M:n}]}_{\text{asymptotic entropy rate}} \right) \\
&= \lim_{N \rightarrow \infty} \sum_{n=1}^N \left(h_{n} - h \right), \\
\end{align}
as claimed.

\end{proof}

\subsection{Predictive information for a finite-state source}\label{sec:finite-state}

The following result shows that predictive information is bounded at a constant when a language puts symbols in an order that respects the correlational structure of the source distribution, when the source distribution has the form of a Hidden Markov Model. On the other hand, we will show in Section~\ref{sec:EE-permutation} that random orders have average predictive information that grows linearly with the sequence length.

\begin{theorem}
    Let $(S_t)_{t\geq 0}$ be a Hidden Markov Model (HMM) with finite state space $\mathcal{S}$ and finite emission alphabet $\mathcal{A}$ generating a bi-infinite stationary stochastic process $\dots, X_{-1}, X_0, X_1, \dots$.
    Let $L \in \mathbb{N}$, and consider the length-$L$ language given by $X_1 \dots X_L$.
 The predictive information is bounded independently of the sequence length $L$:
\begin{equation}
\frac{1}{L} \sum_{i=1}^L \mathrm{I}\left[X_{1\dots i} : X_{i+1 \dots L}\right] = O(1)
\end{equation}
where $O(1)$ contains constants depending on the HMM but not $L$.
\end{theorem}

\begin{proof}
Let $s_i \in \mathcal{S}$ be the state of the HMM after generating $\dots X_{i-2} X_{i-1} X_i$. Note that $s_i$ is a random variable with $\mathrm{H}[s_i] \leq \log |\mathcal{S}|$. Further, $\mathrm{I}[X_{1\dots i} : X_{i+1 \dots L} | s_i] = 0$.
Hence, by the Data Processing Inequality, $\mathrm{I}[X_{1\dots i} : X_{i+1 \dots L}] \leq \mathrm{H}[s_i] \leq \log |\mathcal{S}| = O(1)$ independently of $L$.
\end{proof}

\subsection{Length-2 languages}
We now analyze the most basic case of a code that minimizes predictive information, one in which every meaning is expressed in a string of length 2. We find that a code which minimizes predictive information in this setting performs Independent Components Analysis on the source distribution, with the two characters of the output string representing the two maximally independent factors of the source.

Let $\mathcal{M}$ be a set of meanings with source distribution $p_M$, $\Sigma_1$ and $\Sigma_2$ be disjoint sets of symbols, and $\mathcal{L}$ be a set of languages defined as bijections $L : \mathcal{M} \rightarrow \Sigma_1 \times \Sigma_2$. The predictive information of a language $E(L)$ is the predictive information of the stream of symbols generated by repeatedly sampling meanings from $p_M$, translating them to strings as $s = L(m)$, and concatenating the resulting strings with a delimiter $\# \notin \Sigma_1, \notin \Sigma_2$ between them.

\begin{theorem}
Any language $L^* \in \mathcal{L}$ that achieves $E(L^*) = \mathop{\min}_{L \in \mathcal{L}} E(L)$ has the form
\begin{equation}
L^*(m) = \ell_1(m) \cdot \ell_2(m),
\end{equation}
where $\ell_i$ denotes some mapping $\ell_i : \mathcal{M} \rightarrow \Sigma_i$ and where the outputs from $\ell_1$ and $\ell_2$ have minimal mutual information:
\begin{equation}
\ell_1, \ell_2 = \arg \min \operatorname{I}[\ell_1(M) : \ell_2(M)],
\end{equation}
with the minimization performed over all mappings $\mathcal{M} \rightarrow \Sigma_i$.
\end{theorem}
\begin{proof}
Because the languages have strings of length 2, we calculate predictive information as
\begin{equation}
E = h_1 + h_2 + h_3 - 3h,
\end{equation}
up to length 3, accounting for the delimiter $\#$ attached after the end of the string. The entropy rate $h = \frac{1}{3}\operatorname{H}[M]$ is constant across all languages because they are all bijections, so we ignore the entropy rate going forward. Furthermore, there is no decrease in $n$-gram entropy rate for $n>3$, so we have $h_3 = h$. Dropping all irrelevant constants, $E$ is thus 
\begin{equation}
E \sim h_1 + h_2.
\end{equation}

Calculation of $h_1$: The unigram entropy rate is the entropy of the distribution over symbols generated by first sampling a time index $t$ relative to the most recent delimiter, and then looking at the symbol at that position. For a code of length $T$ (including the delimiter to the right), this is
\begin{align*}
h_1 &= -\sum_{t=1}^T p(t) \sum_{x \in \Sigma_t} p(X_t=x) \log p(t)p(X_t=x) \\
&= -\frac{1}{T} \sum_{t=1}^T \sum_{x \in \Sigma_t} p(X_t=x) \log \frac{1}{T} p(X_t=x) \\
&= -\frac{1}{T} \sum_{t=1}^T \log \frac{1}{T} - \frac{1}{T}\sum_{t=1}^T \sum_{x \in \Sigma_t} p(X_t=x) \log p(X_t=x) \\
&= \log T + \frac{1}{T} \sum_{t=1}^T \operatorname{H}[X_t],
\end{align*}
that is, a constant reflecting how much information is contained in each symbol about its position in the string, plus the average entropy of symbols found in each position. Ignoring constants not affected by the choice of language $L$, in our case with $T=3$ this is
\begin{align}
h_1 \sim \operatorname{H}[X_1] + \operatorname{H}[X_2] + \underbrace{\cancel{\operatorname{H}[X_3]}}_{=0},
\end{align}
where $\operatorname{H}[X_3]=0$ because we always have $X_3=\#$.

Calculation of $h_2$: The bigram entropy rate $h_2$ can be calculated following the same logic, yielding
\begin{equation}
h_2 \sim \underbrace{\operatorname{H}[X_1 \mid X_0]}_{=\operatorname{H}[X_1]} + \operatorname{H}[X_2 \mid X_1] + \underbrace{\cancel{\operatorname{H}[X_3 \mid X_2]}}_{=0},
\end{equation}
where $H[X_1 \mid X_0] = H[X_1]$ because $X_0$ is the left delimiter, which is uninformative about the value of $X_1$. 

Putting these together and ignoring irrelevant constants yields
\begin{align}
E &\sim h_1 + h_2\\
&\sim \operatorname{H}[X_1] + \operatorname{H[X_2]} + \operatorname{H}[X_1] + \operatorname{H}[X_2 \mid X_1] \\
&= \operatorname{H}[X_1] + \operatorname{H}[X_2 \mid X_1] + \operatorname{I}[X_1 : X_2] + \operatorname{H}[X_1] + \operatorname{H}[X_2 \mid X_1] \\
&= 2\operatorname{H}[X_1] + 2\operatorname{H}[X_2 \mid X_1] + \operatorname{I}[X_1 : X_2] \\
&= 2\operatorname{H}[X_1, X_2] + \operatorname{I}[X_1 : X_2] \\
&= 2\operatorname{H}[M] + \operatorname{I}[X_1 : X_2].
\end{align}
Thus, we are left with
\begin{equation}
\label{eq:mi2}
E \sim \operatorname{I}[X_1 : X_2],
\end{equation}
where all remaining constants do not depend on the choice of language $L$. Without loss of generality, we can write $X_1 = \ell_1(M)$ and $X_2 = \ell_2(M)$ for any language $L$ with the appropriate choice of the $\ell_1, \ell_2$, and thus we have that minimal predictive information is achieved by finding functions $\ell_1, \ell_2$ to minimize mutual information:
\begin{equation}
\arg\min \operatorname{I}[\ell_1(M) : \ell_2(M)].
\end{equation}
\end{proof}

\textbf{Remark.} As predictive information is symmetrical with respect to time reversal, the solutions here are symmetric with respect to swapping $\ell_1$ and $\ell_2$.

\textbf{Remark.} The argument reveals that there is a degenerate solution when $|\Sigma_i| \ge |\mathcal{M}|$: you could encode the source $M$ entirely with $\ell_i$, with the other $\ell_{j \neq i}$ a constant function. In that case it is always possible to achieve $\operatorname{I}[\ell_1(M) : \ell_2(M)] = 0$. This result mirrors the claim from \citet{nowak2000evolution} that combinatorial communication requires that the number of available signals is less than the number of available meanings.

\subsection{Length-3 languages}

We now consider codes consisting of strings of length 3. We find that, in this setting, the \emph{order} of the characters in the string is determined by information locality: the non-adjacent characters should be maximally uncorrelated, while the adjacent characters may be more correlated.

Now consider bijective languages $L : \mathcal{M} \rightarrow \Sigma_1 \times \Sigma_2 \times \Sigma_3$ producing strings of length 3, with the alphabets $\Sigma_i$ all disjoint. Now we no longer have invariance with respect to interchanging the features $\ell_1, \ell_2, \ell_3$: the order in which features are expressed now matters. Below, we show that languages which minimize $E$ order these features so as to minimize the mutual information of the nonlocal features $\ell_1$ and $\ell_3$.

\begin{theorem}
Any length-3 language $L^* \in \mathcal{L}$ that achieves $E(L^*) = \mathop{\min}_{L \in \mathcal{L}} E(L)$ has the form
\begin{equation}
L^*(m) = \ell_1(m) \cdot \ell_2(m) \cdot \ell_3(m)
\end{equation}
where the functions $\{\ell_i\}$ are ordered so that $\operatorname{I}[\ell_1(M) : \ell_3(M)]$ is minimal.
\end{theorem}
\begin{proof}
Dropping irrelevant constants in the length-3 case yields
\begin{equation}
E \sim \operatorname{I}[X_1 : X_2] + \operatorname{I}[X_2 : X_3] + 2\operatorname{I}[X_1 : X_3 \mid X_2].
\end{equation}
This expression can be written out and then rearranged as so:
\begin{align}
E &\sim \mathop\mathbb{E}\left[\ln \frac{p(X_1,X_2)}{p(X_1)p(X_2)}\right] + 
\mathop\mathbb{E}\left[\ln \frac{p(X_2,X_3)}{p(X_2)p(X_3)}\right] + 
2\mathop\mathbb{E}\left[\ln \frac{p(X_1,X_2,X_3)p(X_2)}{p(X_1,X_2)p(X_2,X_3)}\right] \\
&= \mathop\mathbb{E}\left[\ln \frac{p(X_1, X_2, X_3)p(X_1,X_2,X_3)p(X_2)p(X_2)}{p(X_1)p(X_2)p(X_3)p(X_1,X_2)p(X_2,X_3)}\right] \\
&= \mathop\mathbb{E}\left[\ln \frac{p(X_1, X_2, X_3)}{p(X_1)p(X_2)p(X_3)}\right] + \mathop\mathbb{E}\left[\ln \frac{p(X_1, X_3 \mid X_2)}{p(X_1 \mid X_2)p(X_3 \mid X_2)}\right] \\
&= \operatorname{TC}[X_1 : X_2 : X_3] + \operatorname{I}[X_1 : X_3 \mid X_2] \\
&= \underbrace{\operatorname{TC}[X_1 : X_2 : X_3] - \operatorname{I}[X_1 : X_2 : X_3]}_{\text{Order-independent}} + \underbrace{\operatorname{I}[X_1 : X_3]}_{\text{Order-dependent}},
\end{align}
where $\operatorname{TC}[\cdot : \cdot : \cdot]$ is total correlation \citep{watanabe1960information} and $\operatorname{I}[\cdot : \cdot : \cdot]$ is multivariate mutual information \citep{mcgill1955multivariate}. Both the TC term and the multivariate mutual information term are invariant to permutations, so the ordering of $X_1,X_2,X_3$ does not matter for them. The only term that depends on the order of symbols is $\operatorname{I}[X_1 : X_3]$. Thus any candidate optimal language $L$ may be improved by permuting the functions $\ell_1, \ell_2, \ell_3$ to minimize $I[\ell_1(M) : \ell_3(M)]$.
\end{proof}

\textbf{Remark.} The multivariate information term $\operatorname{I}[X_1 : X_2 : X_3]$ may be positive or negative. If it is positive, the situation is called redundancy. If it is negative, the situation is called synergy. The result above shows that codes with synergy among the three symbols $X_1, X_2, X_3$ are dispreferred, and codes with redundancy are preferred.

\subsection{Length-$T$ languages}

Next, we consider the more general case of languages with utterance of a fixed length $T$, maintaining the setting where each position in the string has symbols from disjoint alphabets. We show that the predictive information for these languages may be expressed in terms of the coinformation lattice \citep{bell2003coinformation} among random variables corresponding to positions in the string. We find that predictive information is a function of the amount of coinformation in sets of variables and the \emph{span size} of those sets, defined as the linear distance from the first character to the last character in the set. This gives a generalized form of information locality, where predictive information is low whenever any \emph{set} of characters with high \emph{synergy} are all close to each other.

Before stating the result, it is helpful to review the concept of coinformation. Consider a set of $T$ random variables $X_1, \dots, X_T$, and a set of indices such as, for example, $E = \{2,3,4\}$. Let $X_E$ denote the random variables indexed by the set $E$, for example $X_E = \{X_2, X_3, X_4\}$. The \key{coinformation} among the random variables indexed by $E$ is defined as
\begin{equation}
I_E = -\sum_{F \subseteq E} \left(-1\right)^{|F|} \mathrm{H}[X_F],
\end{equation}
that is, the sum of entropies of all the subsets of $X_E$, weighted by $1$ if the subset is of odd cardinality and $-1$ if the subset of is of even cardinality. For example, for $E = \{2,3,4\}$, the coinformation is
\begin{equation}
I_{2,3,4} = H[X_2] + H[X_3] + H[X_4] - H[X_2, X_3] - H[X_3, X_4] - H[X_2, X_4] + H[X_2, X_3, X_4].
\end{equation}

\begin{figure}
\centering
\includegraphics[width=.4\textwidth]{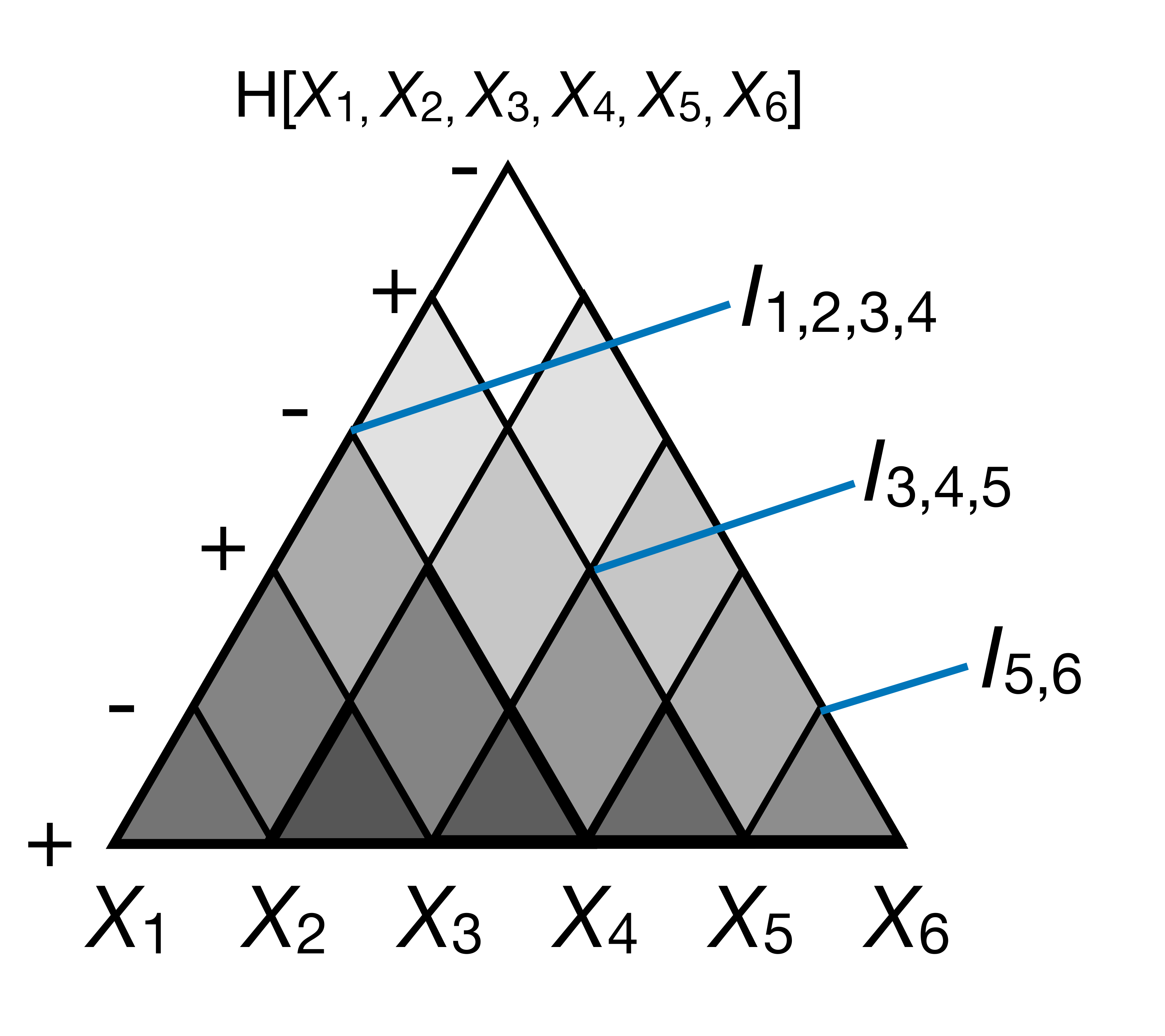}
\label{fig:coinformation-lattice}
\caption{Schematic for coinformation in a set of 6 random variables, based on \citet[][Fig.~2]{bell2003coinformation}. The joint entropy of $X_1, \dots, X_6$ may be found by summing all the coinformations of all the strict subsets of these variables, weighted by the signs given to the left of the triangle. A few coinformations are highlighted. The true lattice of coinformations is a 6D Boolean hypercube; the figure shows a 2D reduction for visual clarity.}
\end{figure}

The coinformation generalizes entropy and mutual information. For a single variable, for example $E = \{1\}$, we recover the univariate entropy, $I_1 = \mathrm{H}[X_1]$. For two random variables, for example $E = \{1,2\}$, we recover mutual information: $I_{1,2} = \mathrm{H}[X_1] + \mathrm{H}[X_2] - \mathrm{H}[X_1, X_2] = \mathrm{I}[X_1 : X_2]$. Coinformation for a set of variables is organized in a lattice structure, as illustrated in Figure~\ref{fig:coinformation-lattice}.

For $|E|$ odd (except for $|E|=1$), coinformation can be negative, which corresponds to \key{synergy}, which happens for three variables when $I[X_1 : X_2 \mid X_3] > I[X_1 : X_2]$. Positive coinformation for an odd number of variables corresponds to \key{redundancy}, which occurs when $I[X_1 : X_2 \mid X_3] < I[X_1 : X_2]$. 

Coinformation may be interpreted as the amount of covariance among variables that cannot be detected from any strict subset of those variables. Synergy reflects a case when there is \emph{more} such covariance, and redundancy reflects a case where there is less. Therefore, we can make the quantity of coinformation somewhat more interpretable by transforming it to \key{synergistic information} $S$, which makes synergy positive and redundancy negative: 
\begin{equation}
\label{eq:syn-info-def}
S_E = \left(-1\right)^{|E|} I_E,
\end{equation}
that is, to define synergistic information, we reverse the sign of coinformation for odd-numbered sets of variables. Synergistic information is positive when there is synergy among an odd-numbered set of variables, negative when there is redundancy, and positive when there is any coinformation among an even-numbered set of variables.

We can now state our result about predictive information in languages consisting of strings of length $T$. 

\begin{theorem}
For a language generating strings of fixed length $T$ and disjoint alphabets for each string position, the predictive information $E$ up to additive and multiplicative constants is
\begin{equation}
E \sim \sum_{1 \le a < \dots < z \le T} \left(z - a \right) S_{a, \dots, z},
\end{equation}
where $S_{a,\dots,z}$ is the synergistic information among the set of random variables $\left\{X_a, \dots, X_z\right\}$ corresponding to characters at positions $a, \dots, z$.
\end{theorem}
\begin{proof}
Up to additive and multiplicative constants, the predictive information in this language is
\begin{align}
E &\sim \sum_{t=1}^T \mathrm{I}[X_1, \dots, X_{t} : X_{t+1}, \dots, X_T] \\ 
&= \sum_{t=1}^T \left( \mathrm{H}[X_1, \dots, X_t] + \mathrm{H}[X_{t+1}, \dots, X_T] - \mathrm{H}[X_1, \dots, X_T] \right).
\end{align}
Next, we note that, inverting the definition of coinformation, we can write the entropy in a set of $N$ variables as \citep[][p. 922]{bell2003coinformation}
\begin{align}
\mathrm{H}[X_1, \dots, X_N] &= -\sum_{1 \le a < \dots < z \le N} \left(-1\right)^{|a, \dots, z|} I_{a, \dots, z} \\
&= -\sum_{1 \le a < \dots < z \le N} S_{a, \dots, z},
\end{align}
where $a, \dots, z$ is a set of indices defining a subset of the variables $X_1, \dots, X_N$. 
We can use this to rewrite the predictive information in terms of synergistic information:
\begin{equation}
E \sim \sum_{t=1}^T \left( -\sum_{1 \le a < \dots < z \le t} S_{a, \dots, z} -\sum_{t+1 \le a < \dots < z \le T} S_{a, \dots, z} + \sum_{1 \le a < \dots < z \le T} S_{a, \dots, z} \right).
\end{equation}
The question now is how many times we are adding in each synergistic information term $S_{a, \dots, z}$ to get the total. We can imagine the whole expression as a sum over cut points $t$ which split the string into two parts, left and right. Within this sum, for each cut point, the last term adds in a synergistic information term $S_{a, \dots, z}$ for each subset of indices $a, \dots, z$, and the first two terms subtract all of the synergistic information terms whose indices are either entirely to the left of the cut point or to its right, leaving only those terms whose indices `straddle' the cut, in the sense that at least one index is $\le t$ and at least one index is $>t$. Thus, we can rewrite predictive information using an indicator variable for whether the set of indices $a, \dots, z$ straddles the cut $t$. Then we count how often this indicator variable is equal to 1, yielding the result:
\begin{align} 
E &\sim \sum_{t=1}^T \sum_{1 \le a < \dots < z \le T} 1_{a \le t < z} S_{a, \dots, z} \\
&= \sum_{1 \le a < \dots < z \le T} \left( \sum_{t=1}^T  1_{a \le t < z}\right) S_{a, \dots, z} \\
&= \sum_{1 \le a < \dots < z \le T} \left(z - a\right) S_{a, \dots, z}.
\end{align}
\end{proof}
\paragraph{Remark.} It can easily be checked that the formula recovers Eq.~\ref{eq:mi2}, which was used in the proof of Theorem~2, for $T=2$.

\paragraph{Remark.} This result goes some way toward linking the hierarchical and well-nested structure of human language with predictive information. In fixed-length languages that minimize predictive information, \emph{groups} of words or letters will tend to be close to each other as a function of how much they covary, in a way that is nested according to the structure of the coinformation lattice. Ill-nested configurations, in which groups of variables with high synergistic information are placed in such a way that other variables intervene, would contribute more to the predictive information, since the synergistic information in groups of variables is weighted by the span of those variables.


\subsection{Predictive information for a random permutation}\label{sec:EE-permutation}

The following result shows that random orders have average predictive information that grows linearly with the sequence length.
This is in contrast to our results from Section~\ref{sec:finite-state} showing that, for finite-state processes, the predictive information is bounded independently of $L$.

\begin{theorem}
    %
    Let $\dots, X_{-1}, X_0, X_1, \dots$ be a bi-infinite stationary process.
    Let $L \in \mathbb{N}$, and consider the length-$L$ language given by $X_1 \dots X_L$.
 Assume the process contains predictive information beyond its ergodic components \citep{dkebowski2009general}, in the sense that:
\begin{equation}
    \inf_{\Delta > 0} \mathrm{I}[X_w : X_{\dots w - \Delta}] < \mathrm{I}[X_w : X_{\dots, w-2, w-1}]
\end{equation}
Consider the uniform distribution over bijections $\rho : [1,\dots,L] \rightarrow [1,\dots,L]$. 
Then
\begin{equation}
\mathbb{E}_\rho\left[\frac{1}{L} \sum_{i=1}^L \mathrm{I}\left[X_{\rho(1\dots i)} : X_{\rho(i+1 \dots L)}\right]\right] = \Theta(L)
\end{equation}
where the expectation describes an average over all bijections $\rho$, and constants in $\Theta(L)$ depend on the HMM but not $L$.
\end{theorem}
The intuition is that for any process with local statistical structure, beyond its ergodic components, permutations of the positions will tend to disrupt this local structure and create long-range dependencies.

\begin{proof}
The expectation is evidently $O(L)$; we need to show it is $\Omega(L)$.
Define $A = \rho(1\dots i)$, $B = \rho(i+1\dots L)$.
The proof idea is to focus attention on positions $w$ where $w \in A$ but a contiguous sequence of positions to its left is in $B$. Such situations create opportunity for $X_B$ to provide predictive information about $X_A$.
Formally, for any $\Delta > 0$:
    \begin{align*}
     & \mathbb{E}[\mathrm{I}[X_A : X_B]] \\
     &= \mathbb{E}\left[\sum_{w \in A} \mathrm{I}[X_w : X_{j \in B} | X_{j < w, j\in A}]\right] \\
    &= \sum_{w=1}^L \mathbb{E}\left[1_{w \in A} \mathrm{I}[X_w : X_{j \in B} | X_{j < w, j\in A}]\right] \\
        &  \geq \sum_{w=1}^L \mathbb{E}\left[1_{w \in A}1_{[w-\Delta,w-1] \cap A = \emptyset} \mathrm{I}[X_w : X_{j \in B} | X_{j<w, j\in A}]\right] \\
    &  \geq \sum_{w=1}^L \mathbb{E}\left[1_{w \in A}1_{[w-\Delta,w-1] \cap A = \emptyset} \mathrm{I}[X_w : X_{[w-\Delta, w-1]} | X_{j<w, j\in A}]\right] \\
    &  = \sum_{w=1}^L \mathbb{E}\left[1_{w \in A}1_{[w-\Delta,w-1] \cap A = \emptyset} \mathrm{I}[X_w : X_{[w-\Delta, w-1]} | X_{j<w-\Delta, j\in A}]\right] \\
   &  = \sum_{w=1}^L p_\rho(w \in A; [w-\Delta,w-1] \cap A = \emptyset) \mathbb{E}\left[\mathrm{I}[X_w : X_{[w-\Delta, w-1]} | X_{j<w-\Delta, j\in A}] | w \in A, 1_{[w-\Delta,w-1] \cap A = \emptyset} \right] \\
    \end{align*}
We now need to show that, for large $\Delta$,
\begin{equation}\label{eq:information-delta-gap}
   \mathrm{I}[X_i : X_{[w-\Delta, w-1]} | X_{j<w-\Delta, j\in A}]
\end{equation}
is bounded away from 0 uniformly over $A$.
Consider\footnote{Reflecting the general identity
\begin{align*}
     \mathrm{I}[A:B \mid C] &= H[A \mid C] - H[A \mid C,B] \\
     &=  H[A] - H[A \mid B] - H[A] + H[A \mid C] + H[A \mid B] - H[A \mid B,C] \\
     &=  \mathrm{I}[A:B] - \mathrm{I}[A:C] + \mathrm{I}[A:C \mid B]
\end{align*}}
\begin{align*}
     &\mathrm{I}[X_w:X_{[w-\Delta, w-1]} \mid X_{j<w-\Delta, j\in A}] \\
     &= \mathrm{I}[X_w:X_{[w-\Delta, w-1]}] - \mathrm{I}[X_w:X_{j<w-\Delta, j\in A}] + \mathrm{I}[X_w:X_{j<w-\Delta, j\in A} \mid X_{[w-\Delta, w-1]}] \\
     &\geq  \mathrm{I}[X_w:X_{[w-\Delta, w-1]}] - \mathrm{I}[X_w:X_{j<w-\Delta, j\in A}]  \\ 
     & \geq  \mathrm{I}[X_w:X_{[w-\Delta, w-1]}] - \mathrm{I}[X_w:X_{j<w-\Delta}]
\end{align*}
When $\Delta \rightarrow \infty$, the first term converges to $\mathrm{I}[X_w|X_{\dots w-2, w-1}]$. 
By assumption, the difference between this and the second term is strictly greater than zero.
Overall, this shows~(\ref{eq:information-delta-gap}) is bounded strictly away from zero independently of $A$, for some sufficiently large $\Delta$ which we henceforth fix for the given HMM, independently of $L$. Let $C>0$ be this lower bound for (\ref{eq:information-delta-gap}).

    It remains to understand why, assuming $|A|$ and $|B|$ are sufficiently large, $\mathbb{E}[\mathrm{I}[X_A : X_B]]$ is $\Omega(L)$.
    Given the $\Delta$ we have fixed,
    \begin{equation}\label{eq:strip-A-probability}
        p_\rho(w \in A; [w-\Delta,w-1] \cap A = \emptyset) \geq D > 0
    \end{equation} for a constant $D$ independent of $w$, for $L$ sufficiently large, when $0.1 L < |A| < 0.9 L$.
For, in this case, we have
\begin{align*}
    & p_\rho(w \in A; [w-\Delta,w-1] \cap A = \emptyset)\\
    &= p_\rho(w \in A) \cdot \prod_{j=1}^\Delta p_\rho(w-j \in B|w \in A, w-1 \in B, \dots, w-j+1 \in B) \\
    &= \underbrace{p(\rho(w) \leq i)}_{=\frac{i}{L}} \cdot \prod_{j=1}^\Delta \underbrace{p\left(\rho(w-j) > i|\rho(w) \leq i, \rho(w-1) > i, \dots, \rho(w-j+1) > i\right)}_{=\frac{L-i-j+1}{L-j}} \\
    & \geq \frac{i}{L} \cdot \left(\frac{L-i-\Delta}{L}\right)^\Delta \\
    & \geq \frac{1}{10} \left(\frac{0.1L-\Delta}{L}\right)^\Delta \\
    & \geq \frac{1}{10} \cdot \frac{1}{20^\Delta} =: D
\end{align*}
where the last step holds when $L > 20\Delta$.
    Taken together, 
    \begin{equation}
        \mathbb{E}[\mathrm{I}[X_A : X_B]] \geq L\cdot D\cdot C = \Omega(L)
    \end{equation}
    when $0.1 L < |A| < 0.9 L$.
    The claim follows.
\end{proof}
We note that one can strengthen the proof to provide a high-probability bound, showing that \emph{most} permutations $\rho$ satisfy such linear scaling. The reason is that a random permutation, when $|A|$ and $|B|$ are both large, is very likely to satisfy the event described in (\ref{eq:strip-A-probability}) on a constant fraction of positions $w$.

\section{Sources over Two Features}
\label{app:two-sweep}

Simulation results in the main text are based on distributions of the form
\begin{equation}
p(M) = p(M_1) \times p(M_2, M_3)
\end{equation}
for varying levels of correlation between the binary random variables $M_2$ and $M_3$. The main result is that when $M_2$ and $M_3$ have lower mutual information, a systematic code for these features minimizes predictive information, but as mutual information increases, a holistic code is more preferred. Here we complement these results with a more in-depth study of a source distribution over two features of the form $p(M) = p(M_1, M_2)$ for binary random variables $M_1$ and $M_2$, looking at a grid of possible distributions over 4 outcomes. This comprehensive approach allows us to examine the effects of the marginal probabilities for $M_1$ and $M_2$, as well as the effects of different \emph{kinds} of correlations between features on the relative preference for systematic vs. holistic codes.

\begin{figure}
\centering
\includegraphics[width=\textwidth]{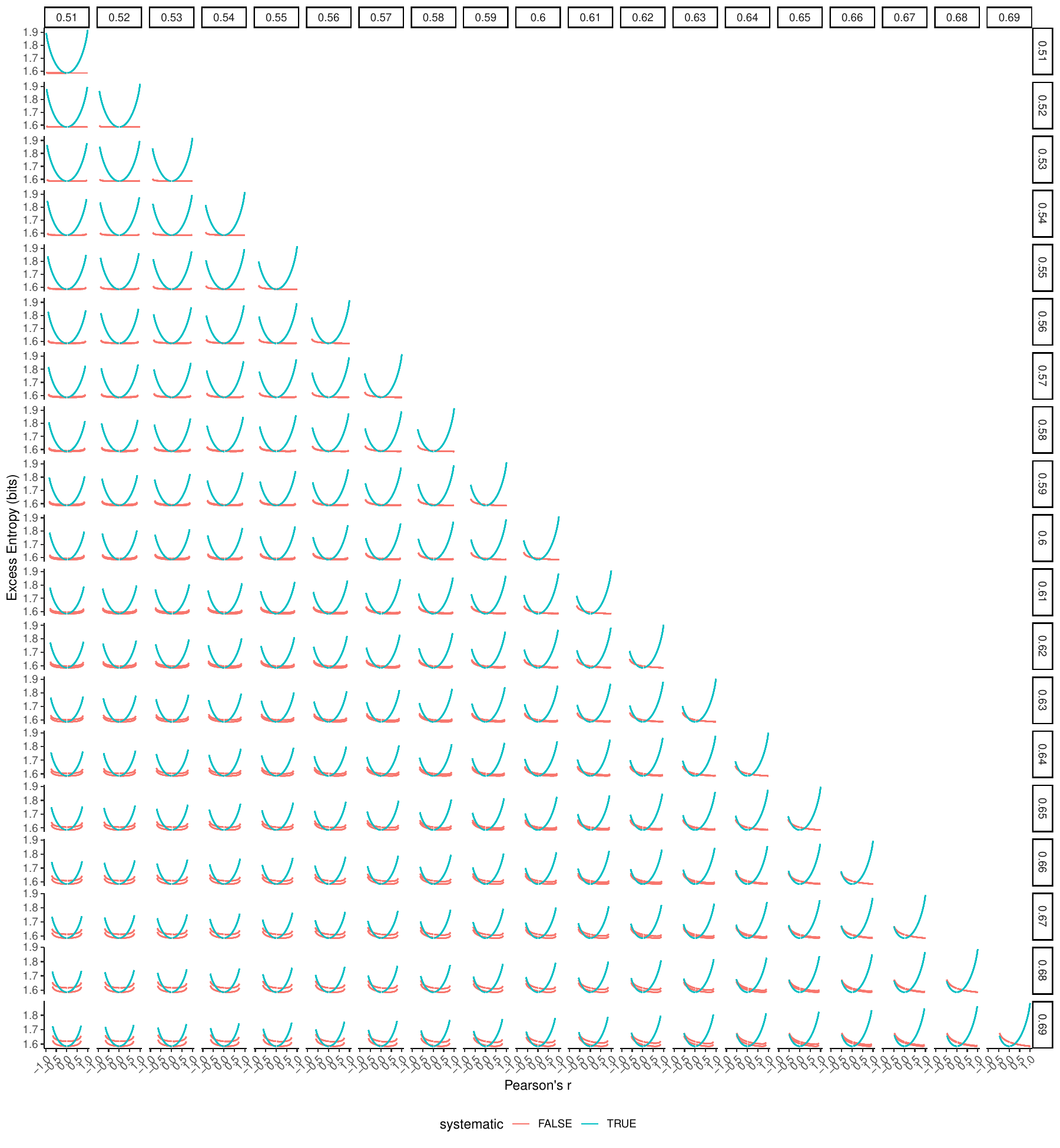}
\caption{Predictive information (labelled as excess entropy) of length-2 codes for a grid over the simplex of possible sources over two binary random variables, $p(M) = p(M_1, M_2)$. Rows show the marginal probability $p(M_1=1)$. Columns show the marginal probability $p(M_2=1)$. The $x$ axis shows the Pearson correlation between $M_1$ and $M_2$.}
\label{fig:two-sweep}
\end{figure}

The main result is shown in Figure~\ref{fig:two-sweep}, which shows predictive information for all possible mappings from the four outcomes of $M$ to strings in $\{\texttt{a},\texttt{b}\}\times \{\texttt{c}, \texttt{d}\}$. The rows indicate different marginal probabilities for $p(M_1=1)$, the columns indicate different marginal probabilities for $p(M_2=1)$, and the $x$ axis indicates the Pearson correlation between $M_1$ and $M_2$. The Pearson correlation is necessary to make sense of the pattern here, because two kinds of correlation can induce mutual information between $M_1$ and $M_2$: a \emph{positive} correlation between the most probable outcomes and a \emph{negative} correlation, as shown in Table~\ref{tab:two-sources}. In the positive correlation case, the features $M_1$ and $M_2$ are effectively `fused'---at maximal correlation, there is actually only one feature here, as we always have $M_1=M_2$. In the negative correlation case, it is as if one of the four outcomes has been effectively removed from the probability distribution.

\begin{table}
\centering
\begin{tabular}{lrrlll}
\toprule
Outcome & Corr. Source & Anticorr. Source & Systematic & cnot(1,2) & cnot(2,1) \\
\midrule
00 & \freqbox{0.375in} \sfrac{3}{8} & 0 & \texttt{ac} & \texttt{ac} & \texttt{ac} \\
01 & \freqbox{0.125in} \sfrac{1}{8} & \freqbox{0.25in} \sfrac{1}{4} & \texttt{ad} & \texttt{ad} & \texttt{bd}  \\
10 & \freqbox{0.125in} \sfrac{1}{8} & \freqbox{0.25in} \sfrac{1}{4} & \texttt{bc} & \texttt{bd} & \texttt{bc} \\
11 & \freqbox{0.375in} \sfrac{3}{8} & \freqbox{0.5in} \sfrac{1}{2} & \texttt{ad} & \texttt{bc} & \texttt{ad} \\
\bottomrule
\end{tabular}
\caption{Some possible sources and codes for the two binary random variables $M_1, M_2$. The correlated source has Pearson's $r=\sfrac{1}{2}$. The anticorrelated source has $r=-\sfrac{1}{3}$.}
\label{tab:two-sources}
\end{table}

There are two conclusions to be drawn from Figure~\ref{fig:two-sweep} beyond the conclusions in the main text. First, the level of preference for systematicity in the low-correlation case depends on the marginal distributions being imbalanced: at $p(M_1)=p(M_2)=\frac{1}{2}$, even when there is zero correlation between the features, the holistic code is just as good as the systematic code. This makes sense because for a uniform distribution over 4 outcomes, there is no reason to favor any one factorization over another. However, as the marginals become more imbalanced (moving downward or to the right in the figure), the systematic code becomes better in the low-correlation range. For these imbalanced marginals, there are generally two red lines to be seen in the figure, corresponding to the two possible classes of non-systematic codes for the source: $\mathrm{cnot}(1,2)$ and $\mathrm{cnot}(2,1)$, which differ in which feature is used as the control bit to flip the other one. 

The second conclusion to be drawn from Figure~\ref{fig:two-sweep} is that there is different behavior for positive and negative feature correlations when the marginals for $M_1$ and $M_2$ are both imbalanced. In particular, in the lower right corner, the systematic code is sometimes better than the holistic code when there is a negative correlation. This happens because, in the negatively correlated source, the systematic code allows the appearance of individual symbols to be correlated with the overall probability of the string: for example, in the systematic code for the negatively correlated source in Table~\ref{tab:two-sources}, high-frequency strings always have $\texttt{d}$, and $\texttt{a}$ only appears in low-frequency strings. The result is that the unigram entropy is minimized by the systematic code for such a source.

\begin{figure}
\centering
\includegraphics[width=\textwidth]{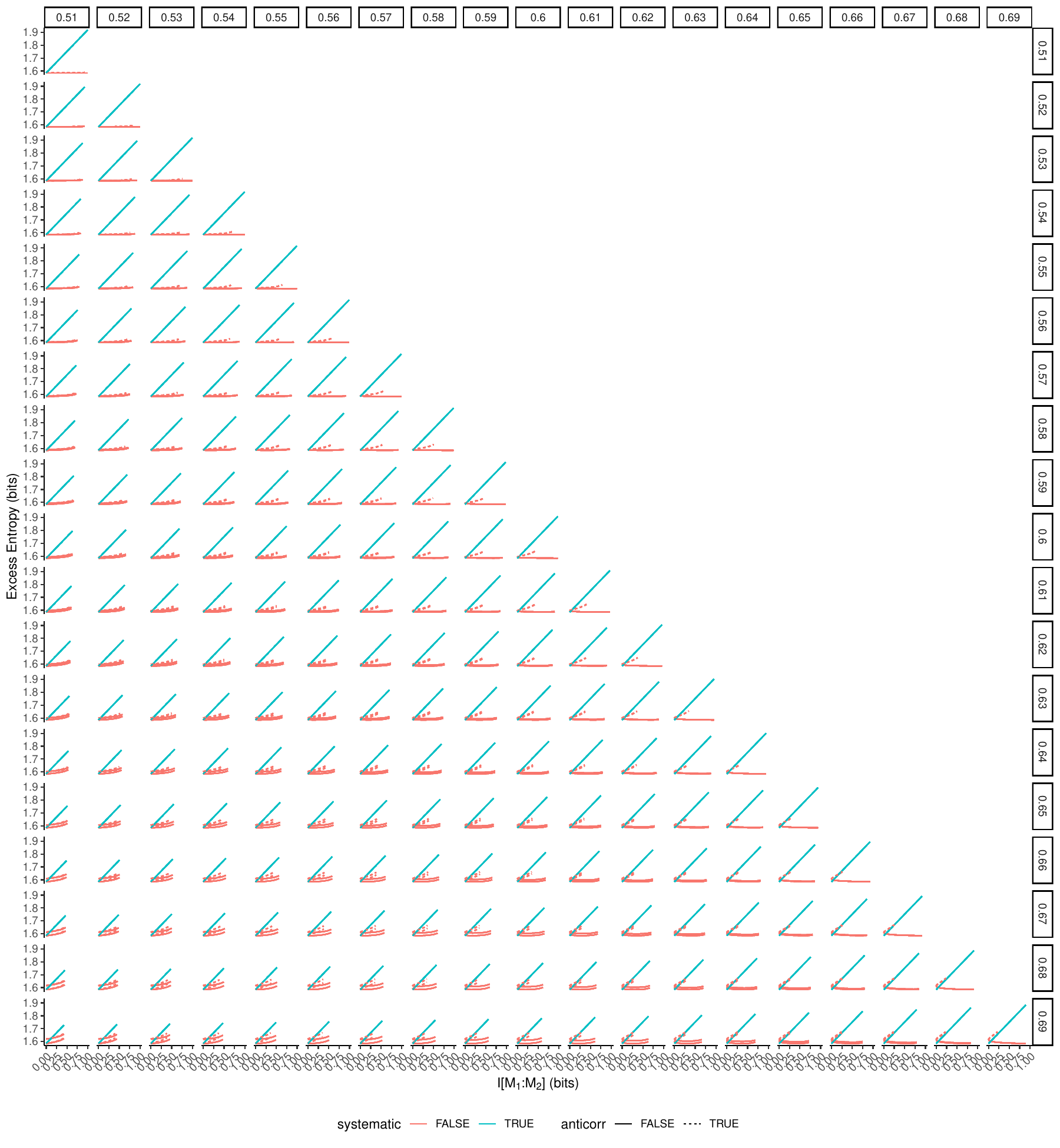}
\caption{Predictive information (labelled as excess entropy) of codes for a grid over the simplex of possible sources over two binary random variables as in Figure~\ref{fig:two-sweep}, but now by mutual information instead of Pearson correlation. Dotted lines indicate codes for sources whose Pearson correlation is negative.}
\label{fig:two-sweep-mi}
\end{figure}

Figure~\ref{fig:two-sweep-mi} shows predictive information for codes as a function of mutual information between random variables $M_1$ and $M_2$, with the negatively-correlated sources separated out and indicated with a dotted line. We see that the preference for holistic codes as a function of mutual information is weaker for the negatively correlated sources, and also that these sources cannot achieve mutual information as high as the positively correlated ones.

\section{Phonological Locality in 61 languages}
\label{app:all-wolex}

\begin{figure}
\centering
\includegraphics[width=\textwidth]{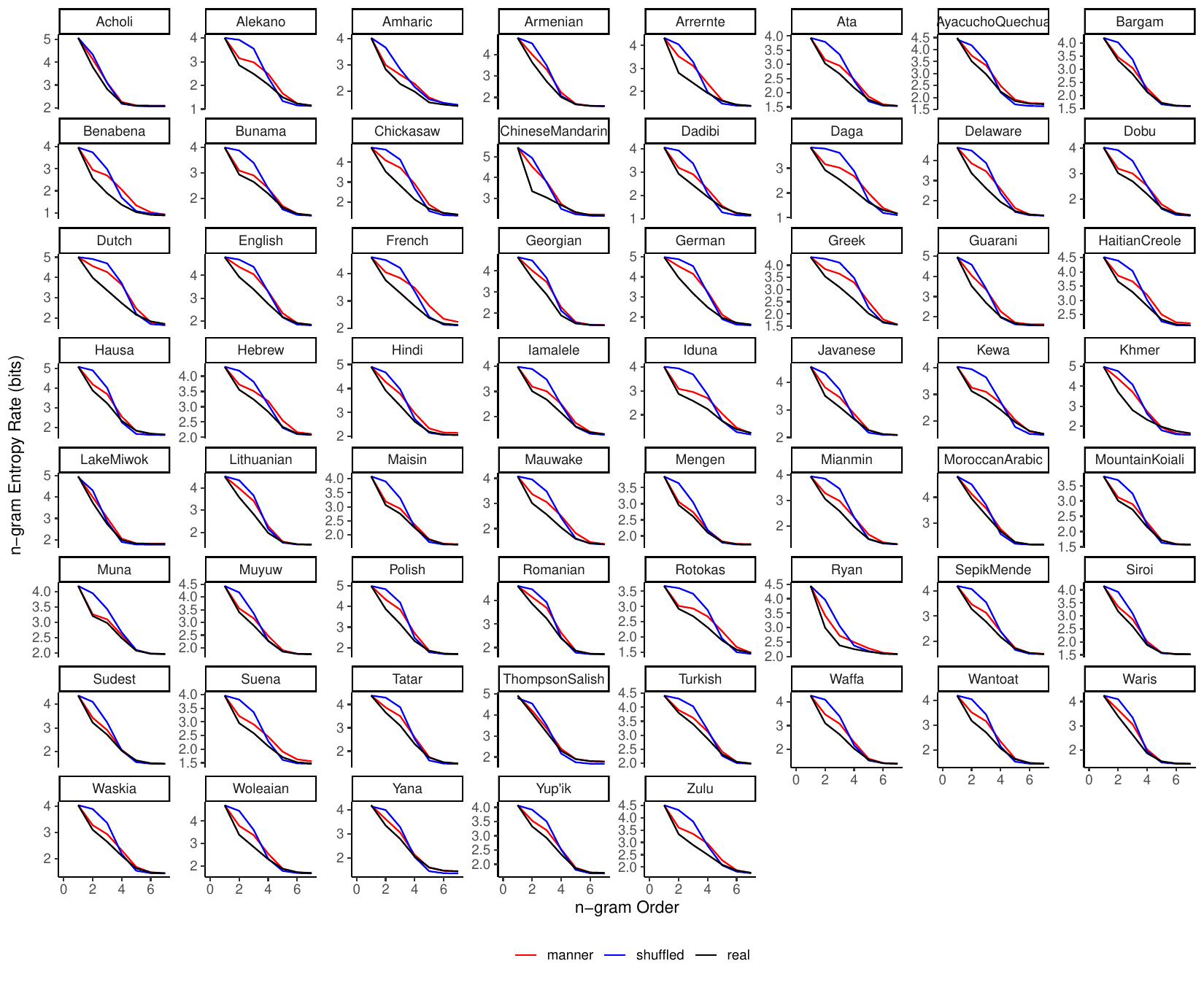}
\caption{Calculation of predictive information for all 61 languages in the WOLEX database, for the attested forms (black), a deterministic shuffle that preserves manner of articulation (red), and a general deterministic shuffle (blue).}
\label{fig:all-wolex}
\end{figure}

Figure~\ref{fig:all-wolex} shows the calculation of predictive information for all 61 languages in the WOLEX database, for the real languages compared against two baselines generated by applying deterministic shuffling functions to the attested forms. Table~\ref{tab:all-wolex} shows the calculated predictive information values. 

\begin{table}
\centering
\tiny{
\begin{tabular}{llll}
\toprule
\textbf{Language} & \textbf{Real} & \textbf{Manner} & \textbf{Shuffled} \\ 
\midrule
Acholi & 5.64 & 6.17 & 6.39 \\
Alekano & 7.42 & 8.79 & 9.60 \\
Amharic & 5.62 & 6.67 & 7.42 \\
Armenian & 6.91 & 8.04 & 8.60 \\
Arrernte & 6.34 & 8.17 & 8.37 \\
Ata & 5.98 & 6.76 & 7.59 \\
Ayacucho Quechua & 6.49 & 7.23 & 7.82 \\
Bargam & 6.26 & 6.74 & 7.45 \\
Benabena & 6.57 & 8.92 & 9.33 \\
Bunama & 6.94 & 7.59 & 8.69 \\
Chickasaw & 8.60 & 10.80 & 11.30 \\
Dadibi & 7.43 & 8.43 & 9.15 \\
Daga & 7.99 & 9.70 & 10.60 \\
Delaware & 7.88 & 9.97 & 10.60 \\
Dobu & 6.99 & 7.86 & 8.91 \\
Dutch & 9.38 & 11.90 & 12.40 \\
English & 6.91 & 8.34 & 8.65 \\
French & 6.35 & 7.78 & 8.50 \\
Georgian & 7.52 & 8.83 & 9.38 \\
German & 8.83 & 11.20 & 11.60 \\
Greek & 8.10 & 10.20 & 10.90 \\
Guarani & 7.00 & 8.21 & 8.66 \\
Haitian Creole & 6.05 & 6.82 & 7.63 \\
Hausa & 8.35 & 9.21 & 9.83 \\
Hebrew & 5.96 & 6.83 & 7.33 \\
Hindi & 6.64 & 7.56 & 8.12 \\
Iamalele & 7.21 & 8.10 & 9.09 \\
Iduna & 8.02 & 9.38 & 10.60 \\
Javanese & 5.72 & 6.57 & 7.08 \\
Kewa & 7.22 & 7.92 & 8.82 \\
Khmer & 8.30 & 10.00 & 10.50 \\
Lake Miwok & 6.20 & 6.75 & 6.87 \\
Lithuanian & 7.24 & 8.54 & 9.00 \\
Maisin & 5.72 & 6.07 & 7.03 \\
Mandarin Chinese & 6.06 & 7.65 & 8.05 \\
Mauwake & 6.53 & 8.05 & 8.79 \\
Mengen & 4.66 & 4.95 & 5.85 \\
Mianmin & 6.80 & 8.01 & 8.83 \\
Moroccan Arabic & 6.20 & 6.71 & 6.99 \\
Mountain Koiali & 5.55 & 5.99 & 6.72 \\
Muna & 5.13 & 5.46 & 6.53 \\
Muyuw & 6.12 & 6.79 & 7.36 \\
Polish & 7.85 & 9.35 & 9.90 \\
Romanian & 7.44 & 8.37 & 8.67 \\
Rotokas & 6.49 & 7.49 & 8.41 \\
Ryan & 3.87 & 5.02 & 5.63 \\
Sepik Mende & 6.77 & 7.48 & 8.47 \\
Siroi & 5.79 & 6.37 & 7.06 \\
Sudest & 6.59 & 6.96 & 7.90 \\
Suena & 6.08 & 6.89 & 7.74 \\
Tatar & 7.96 & 8.82 & 9.39 \\
Thompson Salish & 7.49 & 7.87 & 8.38 \\
Turkish & 7.00 & 7.66 & 8.37 \\
Waffa & 6.64 & 7.74 & 8.49 \\
Wantoat & 6.63 & 7.69 & 8.17 \\
Waris & 6.55 & 7.39 & 7.98 \\
Waskia & 6.40 & 7.09 & 7.85 \\
Woleaian & 6.83 & 7.92 & 8.57 \\
Yana & 6.82 & 7.38 & 8.13 \\
Yup'ik & 6.09 & 6.75 & 7.43 \\
Zulu & 6.79 & 8.12 & 9.00 \\
\bottomrule
\end{tabular}
}
\caption{Predictive information values (in bits) for 61 languages of the WOLEX sample, visualized in Figure~\ref{fig:all-wolex}. `Real' is the predictive information of the attested wordforms. `Manner' is for wordforms shuffled while preserving manner. `Shuffled' is for wordforms shuffled without regard for manner.}
\label{tab:all-wolex}
\end{table}

\section{NP Orders with Other Source Distributions}
\label{app:np-order}

\begin{figure}
\centering
\includegraphics[width=.7\textwidth]{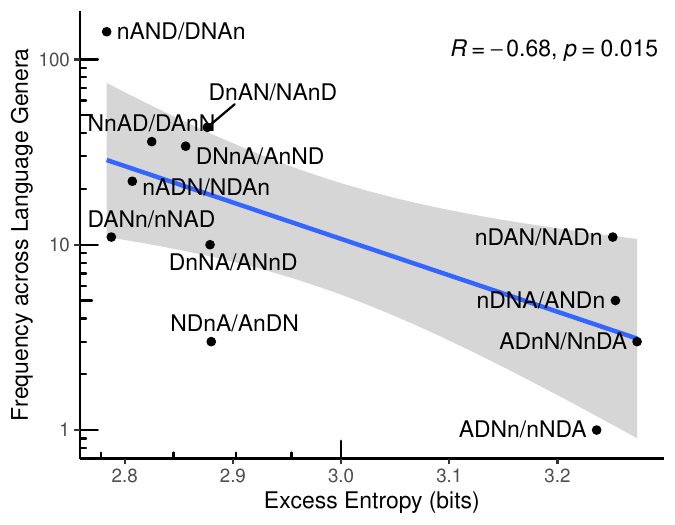}
\caption{Typology frequencies of NP orders by predictive information estimated using the \textbf{Spanish UD source} \citep{taule2008ancora}. Lines and statistics as in the figure in the main text.}
\label{fig:spanish-np}
\end{figure}

\begin{figure}
\centering
\includegraphics[width=.7\textwidth]{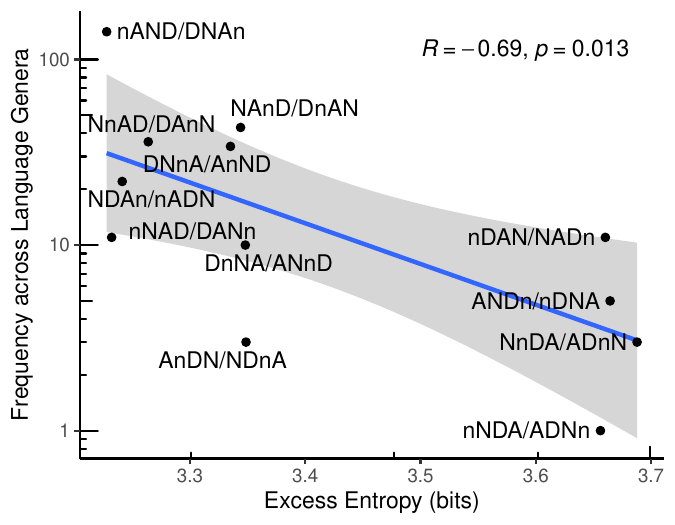}
\caption{Typology frequencies of NP orders by predictive information estimated using the \textbf{English UD source} \citep{zeldes2017gum}.}
\label{fig:english-np}
\end{figure}

\begin{figure}
\centering
\includegraphics[width=.7\textwidth]{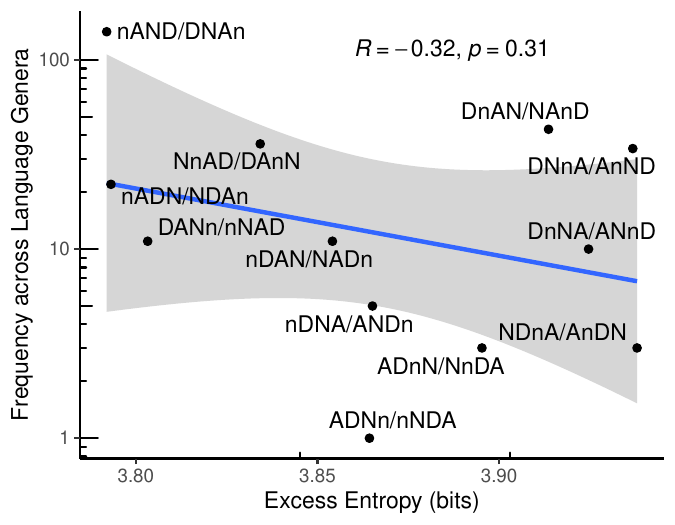}
\caption{Typology frequencies of NP orders by predictive information estimated using the \textbf{Czech UD source} \citep{hladka2008czech}. We believe the weaker correlation here is due to the rarity of determiners in the Czech corpus.}
\label{fig:czech-np}
\end{figure}

\begin{figure}
\centering
\includegraphics[width=.7\textwidth]{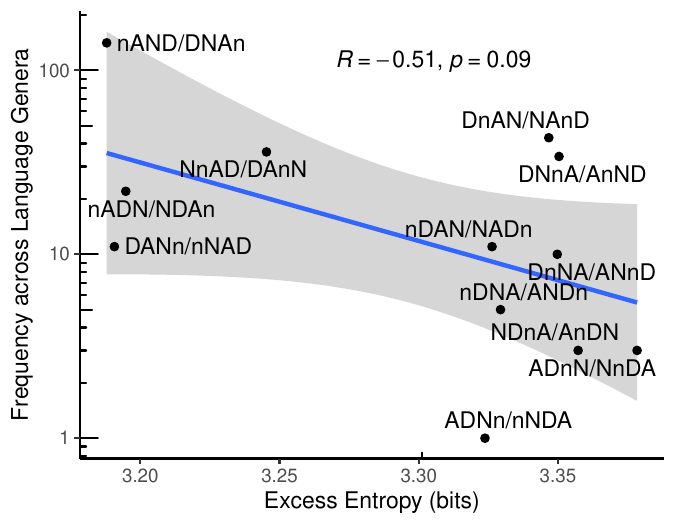}
\caption{Typology frequencies of NP orders by predictive information estimated using the \textbf{Icelandic UD source} \citep{arnardottir2020universal}.}
\label{fig:icelandic-np}
\end{figure}

\begin{figure}
\centering
\includegraphics[width=.7\textwidth]{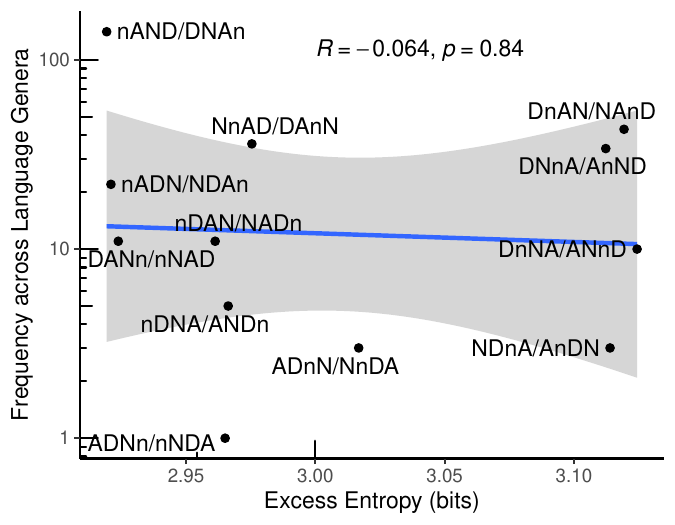}
\caption{Typology frequencies of NP orders by predictive information estimated using the \textbf{Latin UD source} (based on all Latin UD corpora). As the text genre for this corpus is highly unusual (consisting of over 1000 years' worth of text, much of it poetry or written by non-native speakers), we believe that the distribution of NPs in this corpus is not representative of the `true' source distribution over NP meanings.}
\label{fig:latin-np}
\end{figure}

\begin{figure}
\centering
\includegraphics[width=.7\textwidth]{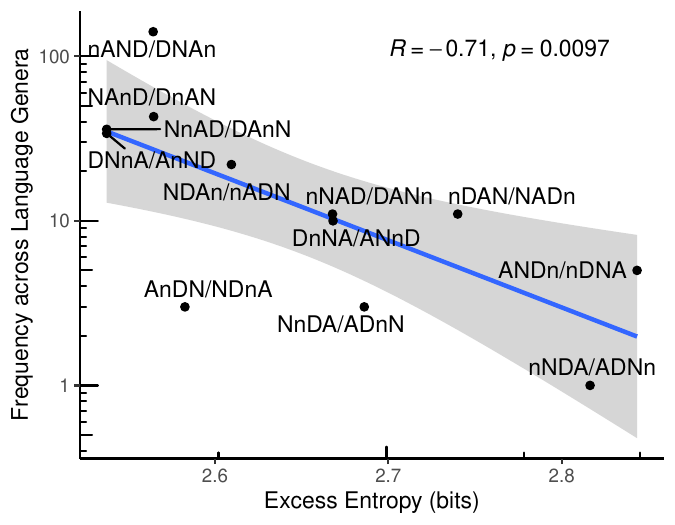}
\caption{Typology frequencies of NP orders by predictive information estimated using the artificial \textbf{MK23 source} \citep{mansfield2023emergence}.}
\label{fig:mksource}
\end{figure}

The noun phrase ordering results in the main text were derived using a source distribution over NPs estimated from the German Universal Dependencies corpus. Here we show results using other naturalistic source distributions, from corpora of  Spanish (Figure~\ref{fig:spanish-np}), English (Figure~\ref{fig:english-np}), Czech (Figure~\ref{fig:czech-np}), Icelandic (Figure~\ref{fig:icelandic-np}), and Latin (Figure~\ref{fig:latin-np}). We also show results using the artificial source developed by \citet{mansfield2023emergence} to study NP order in Figure~\ref{fig:mksource}. 

\section{Correlation of Semantic Features}
\label{app:other-norms}

\begin{figure}
\centering
\includegraphics[width=\textwidth]{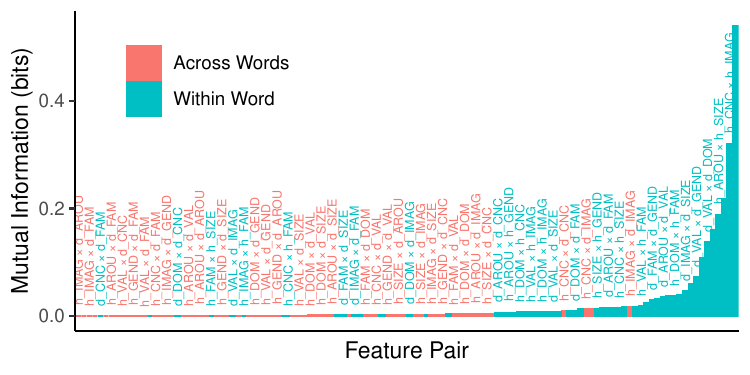} \\
\includegraphics[width=\textwidth]{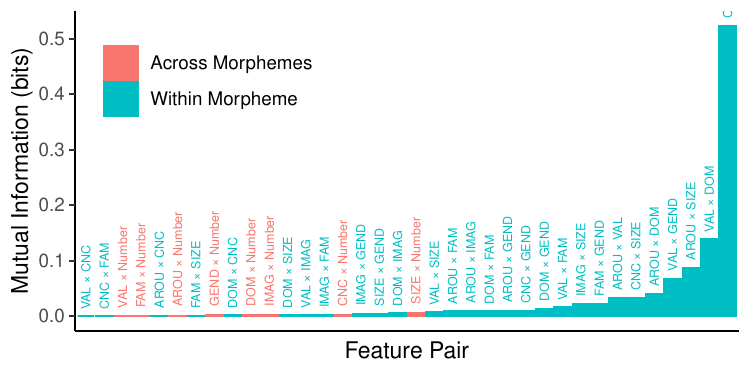} 
\caption{(A) Pairwise mutual information of semantic features from the Glasgow Word Norms \citep{scott2017glasgow} across words and within words, for pairs of verbs and objects in Universal Dependencies English corpora. (B) Pairwise mutual information of the Glasgow Word Norms along with a number feature indicated by plural morphology. The across-word and across-morpheme features have generally lower MI than the within-word and within-morpheme features.}
\label{fig:features2}
\end{figure}

In Figure~\ref{fig:features2} we present results of the study on correlation of semantic features, but using the semantic feature norms from the Glasgow Word Norms \citep{scott2017glasgow}, which rate words for features such as dominance, valence, and arousal. Features are binarized and their frequencies and pairwise MIs are calculated as in the main text. Results are similar to the main text: the across-morpheme and across-word features largely have lower mutual information than within-morpheme and within-word features.

\section{Phonotactic Results with Corpus Frequencies}

\begin{figure}
    \centering\includegraphics[width=0.9\textwidth]{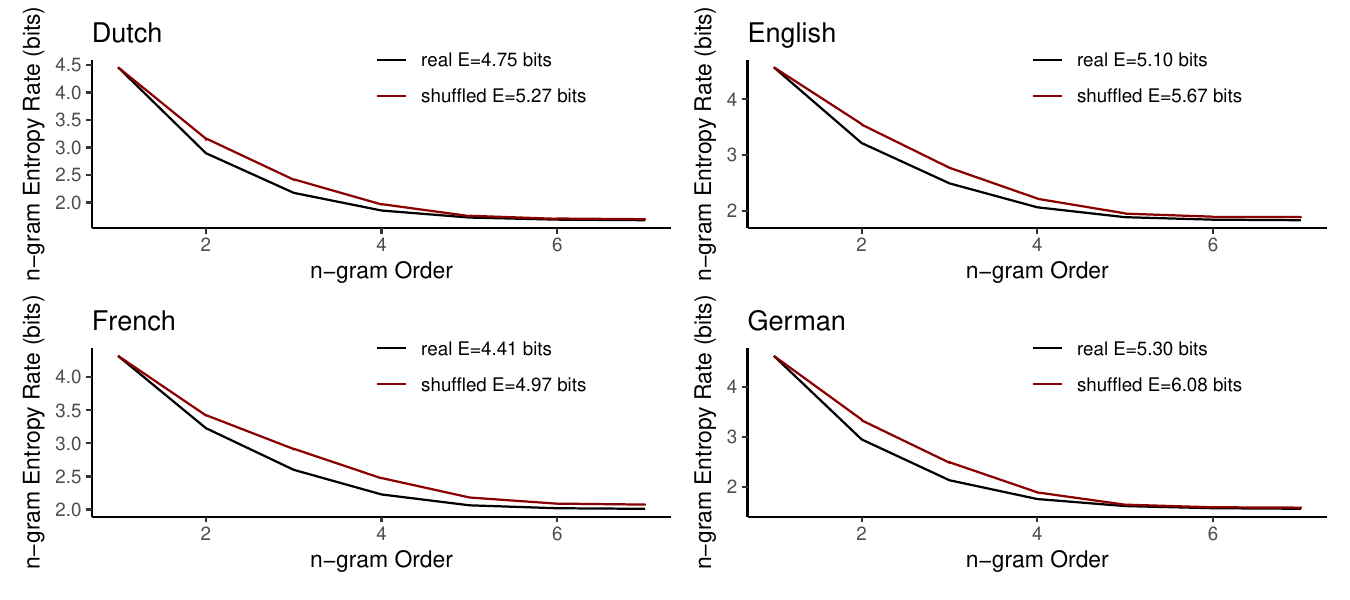}
    \caption{Calculation of predictive information using corpus frequencies, for the 4 languages in the WOLEX database for which orthographic forms are available in WOLEX.
    We show the attested forms (black) and a deterministic shuffle that preserves manner of articulation (red).
    Results match those found with uniform distributions.}
    \label{fig:wolex-with-frequencies}
\end{figure}

Recall that our analysis of phonotactics assumed a uniform distribution over forms. This is because the phonological forms, as listed in the WOLEX database, cannot straightforwardly be matched to corpus data.
However, for four languages (Dutch, English, French, and German), WOLEX provides orthographic forms.
Using these, we derived corpus frequencies from the full Wikipedia texts in these languages. 
We applied simple Laplace smoothing at $\alpha=1$.
Results as shown in Figure~\ref{fig:wolex-with-frequencies} closely agree with those derived under a uniform distribution.

\section{Hierarchically-Structured Sources}

\subsection{Varying Coupling Parameters in Tree Structures}

\begin{figure}
\centering
\includegraphics[width=0.6\textwidth]{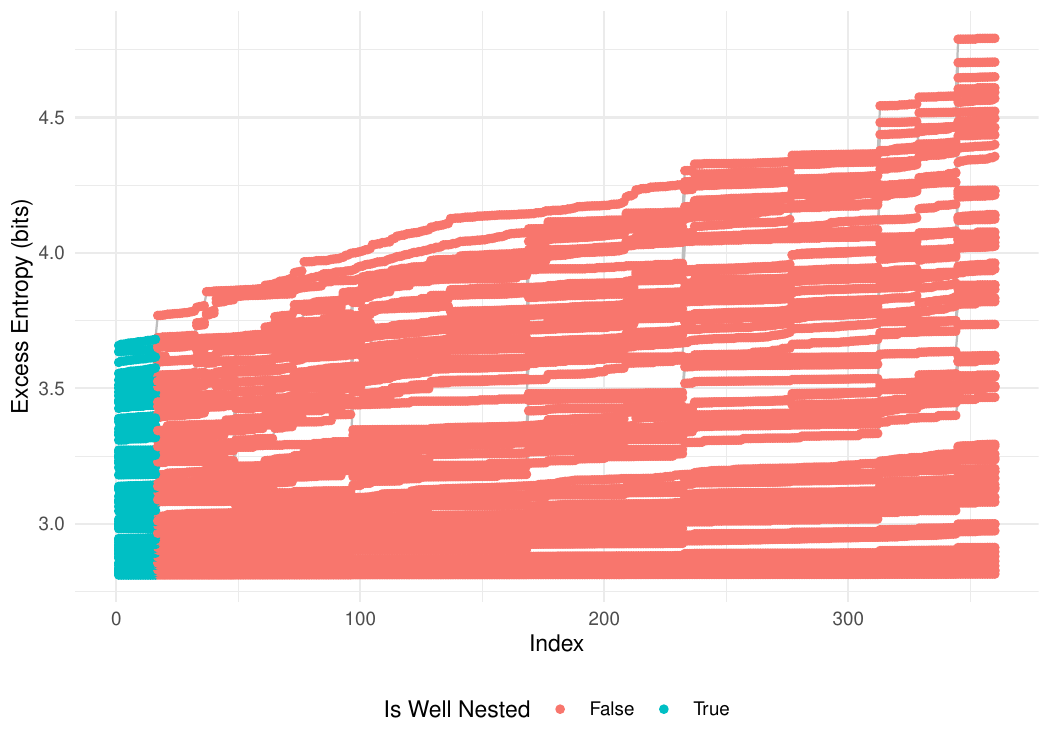}

\caption{Results for 70 sampled combinations of coupling parameters for the tree structure in Main Paper, Figure 2F. Across samples, well-nested orderings achieve lower predictive information than non-well-nested orderings.}
    \label{fig:tree-structures-varied-parameters}
\end{figure}

We created further sources by keeping the tree structure from Main Paper, Figure 2F, but varying the parameters $\alpha, \beta, \gamma \in [0,1]$ randomly subject to the constraint $4 \alpha < 2\beta < \gamma$. We created 70 random samples. Results, shown in Figure~\ref{fig:tree-structures-varied-parameters}, reproduce the pattern from Main Paper, Figure 2F.

\subsection{Sources Defined by PCFGs}

\begin{figure}
\centering
\includegraphics[width=0.99\textwidth]{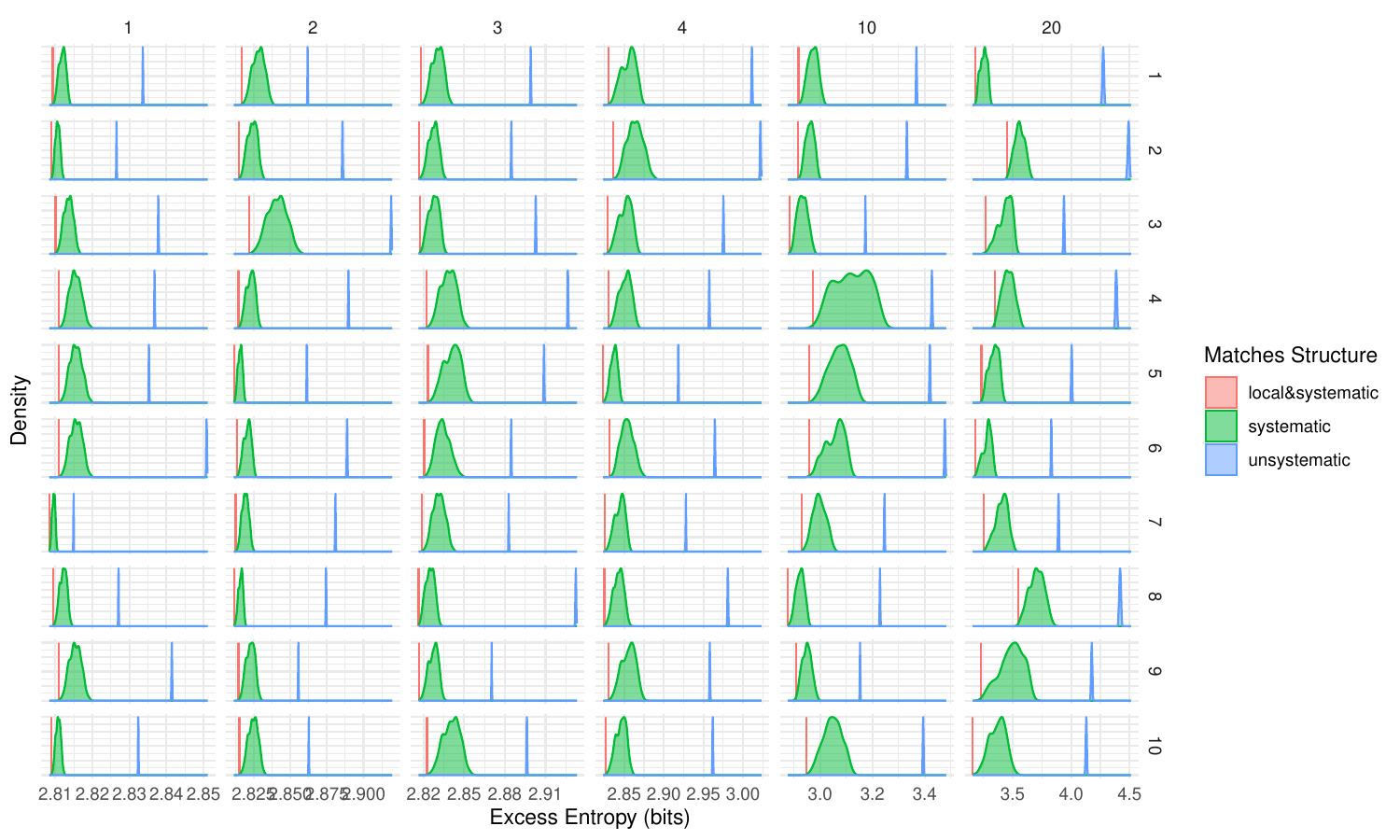}

    \caption{Distribution of predictive information, for 10 randomly constructed PCFG sources for length-6 strings over 5 symbols, at six different inverse temperature parameters ($T$ in (\ref{eq:sampled-pcfg-probabilities})). We compare the local and systematic code given the PCFG (red) with the systematic codes given by the deterministic shuffles of the six positions (green), and an equal number (360, up to reversal) of unsystematic codes given by shuffles of the mapping between forms and probabilities (blue). Local and systematic codes tend to achieve lower predictive information than other systematic codes. Unsystematic codes strongly concentrate at substantially higher predictive information. }
    \label{fig:pcfgs-results}
\end{figure}

We constructed probabilistic context-free grammars (PCFGs) defined by 5 terminals and 5 nonterminals. For each nonterminal $a$, we considered the 100 possible binary productions $a \rightarrow b c$ where $b, c$ are terminals or nonterminals.
For each nonterminal, we defined a distribution over these 100 possible productions $a \rightarrow bc$ by defining
\begin{equation}\label{eq:sampled-pcfg-probabilities}
    p(a \rightarrow bc) \propto \exp(T p_{a \rightarrow bc}),
\end{equation} 
where $T > 0$ is an inverse temperature parameter and each $p_{a \rightarrow bc} \in [0,1]$ is a random number \citep[cf.][]{degiuli2019random}.
The probabilities are normalized to sum up to one for each left-hand side $a$.
The inverse temperature parameter controls the variability in the probabilities of different productions; higher values result in a sparser source.

We then enumerated all $5^6$ strings of length 6 over the given nonterminals, and used the CKY algorithm to compute the probabilities of all of these strings under the given PCFG.
This defines a source over all strings of length 6.

At inverse temperatures $T=1, 2, 3, 4, 10, 20$ we sampled 10 PCFGs each, and compared the predictive information of the language given by the PCFG (systematic and local), deterministic permutations of the 6 positions (systematic and nonlocal), and 360 randomly chosen shuffles of the mapping between forms and probabilities (neither local nor systematic).

Results (Figure~\ref{fig:pcfgs-results}) show that local orderings usually achieve lower predictive information. Nonsystematic codes have much higher predictive information, very closely concentrated around values clearly separated from the systematic codes.

